\pdfoutput=1
\documentclass{article}

\usepackage{microtype}
\usepackage{graphicx}
\usepackage{subfigure}
\usepackage{booktabs} 
\usepackage[utf8]{inputenc} 
\usepackage[T1]{fontenc}    
\usepackage{hyperref}       
\usepackage{url}            
\usepackage{amsfonts}       
\usepackage{nicefrac}       
\usepackage{microtype}      
\usepackage{subfigure}
\usepackage{bbm}
\usepackage{multirow}
\usepackage{amssymb}
\usepackage{float}
\usepackage{xcolor}
\usepackage{algorithm}
\usepackage{algorithmicx}
\usepackage{algpseudocode}
\floatname{algorithm}{Algorithm}

\newcommand*{\fullref}[1]{\hyperref[{#1}]{\autoref*{#1} \nameref*{#1}}}
\usepackage{verbatim}
\usepackage{amsmath}
\usepackage{enumitem}
\usepackage{authblk}

\usepackage{mathtools}
\usepackage{enumitem}
\usepackage{authblk}
\usepackage[numbers,sort]{natbib}
\usepackage[margin=1in]{geometry}
\usepackage{pifont}
\usepackage{ntheorem}
\allowdisplaybreaks[4]
\newtheorem{theorem}{Theorem}

\newtheorem{lemma}{Lemma}
\newtheorem{corollary}{Corollary}
\newtheorem*{proof}{Proof}

\newcommand{\highlight}[1]{\textcolor{purple}{#1}}
\def\*#1{\boldsymbol{#1}}

\newcommand{\R}{\mathbb{R}}
\newcommand{\N}{\mathbb{N}}

\title{Optimal Complexity in Decentralized Training}

\author{Yucheng Lu\thanks{Corresponds to: yl2967@cornell.edu} }
\author{Christopher De Sa\thanks{Corresponds to: cdesa@cs.cornell.edu} }
\affil{Department of Computer Science, Cornell\ University}
\date{}

\begin{document}

\maketitle

\begin{abstract}\label{section abstract}
Decentralization is a promising method of scaling up parallel machine learning systems. In this paper, we provide a tight lower bound on the iteration complexity for such methods in a stochastic non-convex setting. Our lower bound reveals a theoretical gap in known convergence rates of many existing decentralized training algorithms, such as D-PSGD.  We prove by construction this lower bound is tight and achievable. Motivated by our insights, we further propose DeTAG, a practical gossip-style decentralized algorithm that achieves the lower bound with only a logarithm gap. Empirically, we compare DeTAG with other decentralized algorithms on image classification tasks, and we show DeTAG enjoys faster convergence compared to baselines, especially on unshuffled data and in sparse networks.
\end{abstract}
\section{Introduction}\label{section Introduction}
Parallelism is a ubiquitous method to accelerate model training~\cite{abadi2016tensorflow,alistarh2018brief,alistarh2020elastic,lu2020mixml}. 
A parallel learning system usually consists of three layers (Table~\ref{table decentralization}): an application to solve, a communication protocol deciding how parallel workers coordinate, and a network topology determining how workers are connected. 
Traditional design for these layers usually follows a centralized setup: in the application layer, training data is required to be shuffled and shared among parallel workers; while
in the protocol and network layers, workers either communicate via a fault-tolerant single central node (e.g. Parameter Server) \cite{li2014scaling,li2014communication,ho2013more} or a fully-connected topology (e.g. AllReduce) \cite{gropp1999using,patarasuk2009bandwidth}.
This centralized design limits the scalability of learning systems in two aspects. 
First, 
in many scenarios, such as Federated Learning~\cite{koloskova2019decentralized2,mcmahan2016communication} and Internet of Things (IOT)~\cite{kanawaday2017machine}, a shuffled dataset or a complete (bipartite) communication graph is not possible or affordable to obtain.
Second,
a centralized communication protocol can significantly slow down the training, especially with a low-bandwidth or high-latency network \cite{lian2017asynchronous,tang2019doublesqueeze,yu2018distributed}.

\begin{table}[ht!]
\small
\label{table decentralization}
\caption{Design choice of centralization and decentralization in different layers of a parallel machine learning system. The protocol specifies how workers communicate. The topology refers to the overlay network that logically connects all the workers.}
\small
\begin{center}
\begin{tabular}{ccccccc}
\toprule
Layer & Centralized & Decentralized \\
\midrule
\multirow{2}{*}{Application} & \multirow{2}{*}{Shuffled Data} & Unshuffled Data \\
 &  & (Federated Learning) \\
\midrule
\multirow{2}{*}{Protocol} & AllReduce/AllGather & \multirow{2}{*}{Gossip} \\
 & Parameter Server & \\
\midrule
Network & Complete- & \multirow{2}{*}{Arbitrary Graph} \\
Topology & (Bipartite) Graph & \\
\bottomrule
\end{tabular}
\end{center}
\end{table}

\textbf{The rise of decentralization.}
To mitigate these limitations, decentralization comes to the rescue. Decentralizing the application and network allows workers to learn with unshuffled local datasets \cite{li2019convergence} and arbitrary topologies~\cite{seaman2017optimal,shanthamallu2017brief}. Furthermore, the decentralized protocol, i.e. Gossip, helps to balance load, and has been shown to outperform centralized protocols in many cases~\cite{lian2017can,yu2019linear,nazari2019dadam,lu2020moniqua}.

\textbf{Understanding decentralization with layers.}
Many decentralized training designs have been proposed, which can lead to confusion as the term ``decentralization'' is used inconsistently in the literature. Some works use ``decentralized'' to refer to approaches that can tolerate non-iid or unshuffled datasets \cite{li2019convergence}, while others use it to mean gossip communication \cite{lian2017can}, and still others use it to mean a sparse topology graph \cite{wan2020rat}. To eliminate this ambiguity, we formulate Table~\ref{table decentralization}, which summarizes the different ``ways'' a system can be decentralized. Note that the choices to decentralize different layers are \emph{independent}, e.g., the centralized protocol AllReduce can still be implemented on a decentralized topology like the Ring graph \cite{wan2020rat}.

\begin{figure*}[t!]
    \centering
    \includegraphics[width=0.95\linewidth]{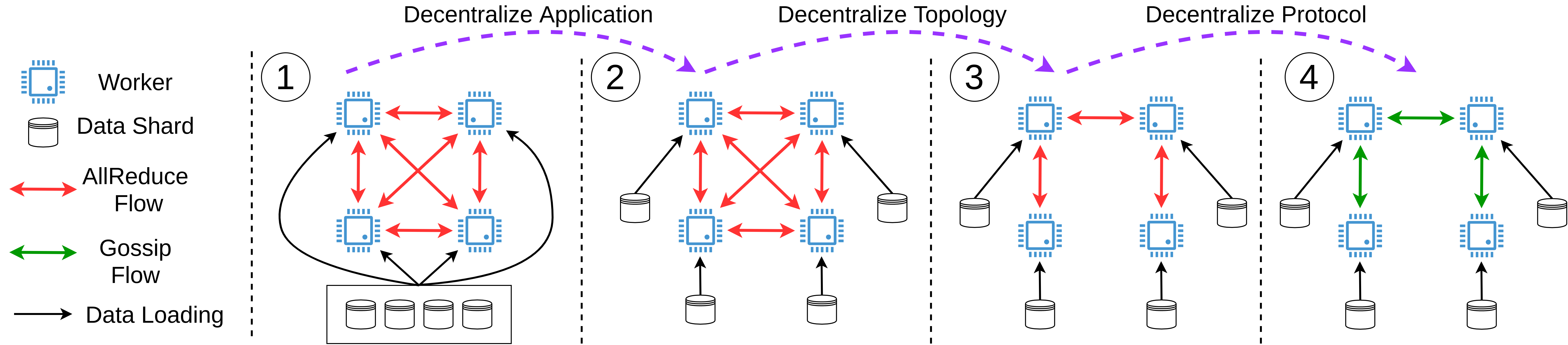}
    \caption{Figure illustrating how decentralization in different layers lead to different learning systems. From left to right: \ding{1}: A fully centralized system where workers sample from shared and shuffled data; \textcircled{2}: Based on \textcircled{1}, workers maintain their own data sources, making it decentralized in the application layer; \textcircled{3}: Based on \textcircled{2}, workers are decentralized in the topology layer; \textcircled{4}: A fully decentralized system in all three layers where the workers communicate via Gossip. Our framework and theory are applicable to all kinds of decentralized learning systems.}
    \label{fig:example}
\end{figure*}

\textbf{The theoretical limits of decentralization.} Despite the empirical success, the best convergence rates achievable by decentralized training---and how they interact with different notions of decentralization---remains an open question.
Previous works often show complexity of a given decentralized algorithm with respect to the number of iterations $T$ or the number of workers $n$, ignoring other factors including network topologies, function parameters or data distribution. Although a series of decentralized algorithms have been proposed showing theoretical improvements---such as using variance reduction~\cite{tang2018d}, acceleration~\cite{seaman2017optimal}, or matching~\cite{wang2019matcha}---we do not know how close they are to an ``optimal'' rate or whether further improvement is possible.

In light of this, a natural question is: \textit{What is the optimal complexity in decentralized training? Has it been achieved by any algorithm yet?} 
Previous works have made initial attempts on this question, by analyzing this theoretical limit in a non-stochastic or (strongly) convex setting~\cite{seaman2017optimal,scaman2018optimal,koloskova2020unified,woodworth2018graph,dvinskikh2019decentralized,sun2019distributed}.
These results provide great heuristics but still leave the central question open, since stochastic methods are usually used in practice and many real-world problems of interest are non-convex (e.g. deep learning).
In this paper we give the first full answer to this question: our contributions are as follows.


\begin{itemize}[nosep,leftmargin=12pt]
    \item In Section~\ref{section limit}, we prove the first (to our knowledge) tight lower bound for decentralized training in a stochastic non-convex setting.  Our results reveal an asymptotic gap between our lower bound and known convergence rates of existing algorithms.
    \item In Section~\ref{section optimal}, we prove our lower bound is tight by exhibiting an algorithm called DeFacto that achieves it---albeit while only being decentralized in the sense of the application and network layers.
    \item In Section~\ref{section detag}, we propose DeTAG, a practical algorithm that achieves the lower bound with only a logarithm gap and that is decentralized in all three layers.
    \item In Section~\ref{section experiment}, we experimentally evaluate DeTAG on the CIFAR benchmark and show it converges faster compared to decentralized learning baselines.
\end{itemize}

\section{Related Work}\label{section related work}
\textbf{Decentralized Training.}
In the application layer, decentralized training usually denotes federated learning \cite{zhao2018federated}. Research on decentralization in this sense investigates convergence where each worker samples only from a local dataset which is not independent and identically distributed to other workers' datasets \cite{bonawitz2019towards,tran2019federated,yang2019federated,konevcny2016federated}. 
Another line of research on decentralization focuses on the protocol layer---with average gossip \cite{boyd2005gossip,boyd2006randomized}, workers communicate by averaging their parameters with neighbors on a graph.
D-PSGD \cite{lian2017can} is one of the most basic algorithms that scales SGD with this protocol, achieving a linear parallel speed up. Additional works extend D-PSGD to asynchronous and variance-reduced cases \cite{lian2017asynchronous,tang2018d,tian2020achieving,zhang2019asyspa,hendrikx2019asynchronous,xin2021hybrid}.
After those, \citet{zhang2019decentralized,xin2019variance,xin2021improved} propose adding gradient trackers to D-PSGD.
Other works discuss the application of decentralization on specific tasks such as linear models or deep learning~\cite{he2018cola,assran2018stochastic}. 
\citet{zhang2019asynchronous} treats the case where only directed communication can be performed. 
\citet{wang2019matcha} proposes using matching algorithms to optimize the gossip protocol.
Multiple works discuss using compression to decrease communication costs in decentralized training~\cite{koloskova2019decentralized,koloskova2019decentralized2,lu2020moniqua,tang2019texttt,tang2018communication}, and other papers connect decentralized training to other parallel methods and present a unified theory~\cite{lu2020mixml,koloskova2020unified,wang2018cooperative}. In some even earlier works like \cite{nedic2009distributed,duchi2010distributed}, full local gradients on a convex setting is investigated.

\textbf{Lower Bounds in Stochastic Optimization.}
Lower bounds are a well studied topic in non-stochastic optimization, especially in convex optimization~\cite{agarwal2014lower,arjevani2015communication,lan2018optimal,fang2018spider,arjevani2017oracle}. 
In the stochastic setting, \citet{allen2018make} and \citet{foster2019complexity} discuss the complexity lower bound to find stationary points on convex problems. 
Other works study the lower bound in a convex, data-parallel setting~\cite{diakonikolas2018lower,balkanski2018parallelization,tran2019adaptive}, and \citet{colin2019theoretical} extends the result to a model-parallel setting.
In the domain of non-convex optimization, \citet{carmon2017lower,carmon2019lower} propose a zero-chain model that obtains tight bound for a first order method to obtain stationary points.
\citet{zhou2019lower} extends this lower bound to a finite sum setting, and \citet{arjevani2019lower} proposes a probabilistic zero-chain model that obtains tight lower bounds for first-order methods on stochastic and non-convex problems.

\begin{table*}[t]
\small
\label{table complexity}
\caption{Complexity comparison among different algorithms in the stochastic non-convex setting on arbitrary graphs.  The \textcolor{blue}{blue} text are the results from this paper. Definitions to all the parameters can be found in Section~\ref{section setting}. Other algorithms like EXTRA \cite{shi2015extra} or MSDA \cite{scaman2017optimal} are not comparable since they are designed for (strongly) convex problems. Additionally, \citet{liu2021distributed} provides alternative complexity bound for algorithms like D-PSGD which improves upon the spectral gap. However, the new bound would compromise the dependency on $\epsilon$, which does not conflict with our comparison here.}
\small
\begin{center}
\begin{tabular}{lllllll}
\toprule
& Source & Protocol & Sample Complexity & Comm. Complexity & Gap to Lower Bound\\
\midrule
\multirow{2}{*}{Lower Bound} & Theorem~\ref{thmlowerbound} & Central & \textcolor{blue}{$\Omega\left(\frac{\Delta L\sigma^2}{nB\epsilon^4}\right)$} & \textcolor{blue}{$\Omega\left(\frac{\Delta LD}{\epsilon^2}\right)$} & /\\

& Corollary~\ref{corollary lower bound} & Decentral & \textcolor{blue}{$\Omega\left(\frac{\Delta L\sigma^2}{nB\epsilon^4}\right)$} & \textcolor{blue}{$\Omega\left(\frac{\Delta L}{\epsilon^2\sqrt{1-\lambda}}\right)$} & /\\
\midrule
\multirow{12}{*}{Upper Bound} & DeFacto (Theorem~\ref{thmrateDeFacto}) & Central & \textcolor{blue}{$O\left(\frac{\Delta L\sigma^2}{nB\epsilon^4}\right)$} & \textcolor{blue}{$O\left(\frac{\Delta LD}{\epsilon^2}\right)$} & \highlight{$O(1)$} \\

& DeTAG (Theorem~\ref{thmrateDeTAG}) & Decentral & \textcolor{blue}{$O\left(\frac{\Delta L\sigma^2}{nB\epsilon^4}\right)$} & $\textcolor{blue}{O\left(\frac{\Delta L\log\left(\frac{\varsigma_0n}{\epsilon\sqrt{\Delta L}}\right)}{\epsilon^2\sqrt{1-\lambda}}\right)}$ & \highlight{$O\left(\log\left(\frac{\varsigma_0n}{\epsilon\sqrt{\Delta L}}\right)\right)$} \\

& D-PSGD \cite{lian2017can} & Decentral & $O\left(\frac{\Delta L\sigma^2}{nB\epsilon^4}\right)$ & $O\left(\frac{\Delta Ln\varsigma}{\epsilon^2(1-\lambda)^2}\right)$ & $O\left(\frac{n\varsigma}{(1-\lambda)^{\frac{3}{2}}}\right)$ \\

& SGP \cite{assran2019stochastic} & Decentral & $O\left(\frac{\Delta L\sigma^2}{nB\epsilon^4}\right)$ & $O\left(\frac{\Delta Ln\varsigma}{\epsilon^2(1-\lambda)^2}\right)$ & $O\left(\frac{n\varsigma}{(1-\lambda)^{\frac{3}{2}}}\right)$ \\

& D$^2$ \cite{tang2018d} & Decentral & $O\left(\frac{\Delta L\sigma^2}{nB\epsilon^4}\right)$ & $O\left(\frac{\lambda^2\Delta Ln\varsigma_0}{\epsilon^2(1-\lambda)^3}\right)$ & $O\left(\frac{\lambda^2n\varsigma_0}{(1-\lambda)^{\frac{5}{2}}}\right)$ \\

& DSGT \cite{zhang2019decentralized} & Decentral & $O\left(\frac{\Delta L\sigma^2}{nB\epsilon^4}\right)$ & $O\left(\frac{\lambda^2\Delta Ln\varsigma_0}{\epsilon^2(1-\lambda)^3}\right)$ & $O\left(\frac{\lambda^2n\varsigma_0}{(1-\lambda)^{\frac{5}{2}}}\right)$ \\

& GT-DSGD \cite{xin2021improved} & Decentral & $O\left(\frac{\Delta L\sigma^2}{nB\epsilon^4}\right)$ & $O\left(\frac{\lambda^2\Delta Ln\varsigma_0}{\epsilon^2(1-\lambda)^3}\right)$ & $O\left(\frac{\lambda^2n\varsigma_0}{(1-\lambda)^{\frac{5}{2}}}\right)$ \\
\bottomrule
\end{tabular}
\end{center}
\end{table*}

\section{Setting}\label{section setting}
In this section, we introduce the notation and assumptions we will use.
Throughout the paper, we consider the standard \emph{data-parallel} training setup with \highlight{$n$} parallel workers.
Each worker $i$ stores a copy of the model $\*x\in\mathbb{R}^d$ and a local dataset $\mathcal{D}_i$. The model copy and local dataset define a local loss function (or empirical risk) $f_i$. The ultimate goal of the parallel workers is to output a target model $\*{\hat{x}}$ that minimizes the average over all the local loss functions, that is,
\begin{equation}\label{objective}
    \*{\hat{x}}=\arg \min_{\*x\in\mathbb{R}^d} \left[ f(\*x) = \frac{1}{n}\sum_{i=1}^{n}\underbrace{\mathbb{E}_{\xi_i\sim\mathcal{D}_i}f_i(\*x;\xi_i)}_{f_i(\*x)} \right].
\end{equation}
Here, $\xi_i$ is a data sample from $\mathcal{D}_i$ and is used to compute a stochastic gradient via some oracle, e.g. back-propagation on a mini-batch of samples.
The loss functions can (potentially) be non-convex so finding a global minimum is NP-Hard; instead,
we expect the workers to output a point $\*{\hat{x}}$ at which $f(\*{\hat{x}})$ has a small gradient magnitude in expectation: \highlight{$\mathbb{E}\|\nabla f(\*{\hat{x}})\|\leq\epsilon$}, for some small $\epsilon$.\footnote{There are many valid stopping criteria. We adopt $\epsilon$-stationary point as the success signal. $\mathbb{E}\|\nabla f(\*{\hat{x}})\|^2\leq\epsilon^2$ is another commonly used criterion; we adopt the non-squared one following \cite{carmon2019lower}. Other criterions regarding stationary points can be converted to hold in our theory.}
The assumptions our theoretical analysis requires can be categorized by the layers from Table~\ref{table decentralization}: in each layer, ``being decentralized'' corresponds to certain assumptions (or lack of assumptions).
We now describe these assumptions for each layer separately.

\subsection{Application Layer}

Application-layer assumptions comprise constraints on the losses $f_i$ from (\ref{objective}) and the gradient oracle via which they are accessed by the learning algorithm, as these are constraints on the learning task itself.

\textbf{Function class ($\Delta$ and $L$).} As is usual in this space, we assume the local loss functions $f_i: \R^d \rightarrow \R$ are $L$-smooth,
\begin{equation}
    \|\nabla f_i(\*x)-\nabla f_i(\*y)\| \leq L\|\*x-\*y\|, \; \forall \*x, \*y\in\mathbb{R}^d,
\end{equation}
for some constant $L > 0$, and that the total loss $f$ is range-bounded by $\Delta$ in the sense that $f(\*0)-\inf_{\*x}f(\*x)\leq\Delta$.
We let the \highlight{function class $\mathcal{F}_{\Delta, L}$} denote the set of all functions that satisfy these conditions (for any dimension $d \in \N^+$).

\textbf{Oracle class ($\sigma^2$).} We assume each worker interacts with its local function $f_i$ only via a stochastic gradient oracle $\tilde{g}_i$, and that when we query this oracle with model $\*x$, it returns an independent unbiased estimator to $\nabla f_i(\*x)$ based on some random variable $z$ with distribution $Z$ (e.g. the index of a mini-batch randomly chosen for backprop). Formally,
\begin{equation}
    \mathbb{E}_{z\sim Z}[\tilde{g}_i(\*x,z)] = \nabla f_i(\*x), \; \forall \*x\in\mathbb{R}^d.
\end{equation}
As per the usual setup, we additionally assume the local estimator has bounded variance: for some constant $\sigma > 0$,
\begin{equation}
    \mathbb{E}_{z\sim Z}\|\tilde{g}_i(\*x,z)-\nabla f_i(\*x)\|^2\leq \sigma^2, \; \forall \*x\in\mathbb{R}^d.
\end{equation}
We let $O$ denote a set of these oracles $\{\tilde{g}_i\}_{i\in[n]}$, and let the \highlight{oracle class $\mathcal{O}_{\sigma^2}$} denote the class of all such oracle sets that satisfy these two assumptions.

\textbf{Data shuffling ($\varsigma^2$ and $\varsigma_0^2$).}
At this point, an analysis with a centralized application layer would make the additional assumption that all the $f_i$ are equal and the $\tilde g_i$ are identically distributed: this roughly corresponds to the assumption that the data all comes independently from a single centralized source.
\emph{We do not make this assumption}, and lacking such an assumption is what makes an analysis decentralized in the application layer.
Still, some assumption that bounds the $f_i$ relative to each other somehow is needed: we now discuss two such assumptions used in the literature, from which we use the weaker (and more decentralized) one. 

One commonly made assumption~\cite{lian2017can,koloskova2019decentralized,koloskova2019decentralized2,lu2020moniqua,tang2018communication} in decentralized training is
\begin{equation}\label{assumeclosedistribution}
    \frac{1}{n}\sum_{i=1}^{n}\|\nabla f_i(\*x) - \nabla f(\*x)\|^2\leq \varsigma^2, \; \forall \*x\in\mathbb{R}^d,
\end{equation}
for some constant $\varsigma$, which is said to bound the ``outer variance'' among workers.
This is often unreasonable, as it suggests the local datasets on workers must have close distribution: in practice, ensuring this often requires some sort of shuffling or common centralized data source. 
We \textbf{do not assume} (\ref{assumeclosedistribution}) but instead adopt the much weaker assumption
\begin{equation}\label{assumeweakclosedistribution}
    \frac{1}{n}\sum_{i=1}^{n}\|\nabla f_i(\*0) - \nabla f(\*0)\|^2\leq \varsigma_0^2,
\end{equation}
for constant $\varsigma_0 > 0$.\footnote{As we only use $\varsigma_0$ for upper bounds, not lower bounds, we do not define a ``class'' that depends on this parameter.}
This assumption only requires a bound at point $\*0$, which is, to the best of our knowledge, the weakest assumption of this type used in the literature \cite{tang2018d,zhang2019decentralized}.
Requiring such a weak assumption allows workers to (potentially) sample from different distributions or vary largely in their loss functions (e.g. in a federated learning environment).

\subsection{Protocol Layer}

Protocol-layer assumptions comprise constraints on the parallel learning algorithm itself, and especially on the way that the several workers communicate to approach consensus.

\textbf{Algorithm class ($B$).}
We consider algorithms $A$ that divide training into multiple iterations, and between two adjacent iterations, there must be a synchronization process among workers (e.g. a barrier) such that they start each iteration simultaneously.\footnote{We consider synchronous algorithms only here for simplicity of presentation; further discussion of extension to asynchronous algorithms is included in the supplementary material.}
Each worker running $A$ has a local copy of the model, and we let $\*x_{t,i} \in \mathbb{R}^d$ denote this model on worker $i$ at iteration $t$.
We assume without loss of generality that $A$ initializes each local model at zero: $\*x_{0,i}=\*0$ for all $i$.
At each iteration, each worker makes \emph{at most} $B$ queries to its gradient oracle $\tilde g_i$, for some constant $B \in \mathbb{N}^+$, and then uses the resulting gradients to update its model.
We do not make any explicit rules for output and allow the output of the algorithm $\*{\hat{x}}_t$ at the end of iteration $t$ (the model that $A$ would output if it were stopped at iteration $t$) to be any linear combination of all the local models, i.e.
\begin{equation}
    \textstyle
    \*{\hat{x}}_t \in \text{span}(\{\*x_{t,j}\}_{j\in[n]}) = \{ \sum_{j=1}^n c_j \*x_{t,j} \mid c_j \in \R \}. 
\end{equation}
Beyond these basic properties, we further require $A$ to satisfy the following ``zero-respecting'' property from \citet{carmon2017lower}.
Specifically, if $\*z$ is any vector worker $i$ queries its gradient oracle with at iteration $t$, then for any $k \in [d]$, if $\*e_k^\top\*z \ne 0$, then there exists a $s \le t$ and a $j \in [n]$ such that either $j = i$ or $j$ is a neighbor of $i$ in the network connectivity graph $G$ (i.e. $(i,j) \in \{(i,i)\} \cup G$) and $(\*e_k^\top\*x_{s,j}) \ne 0$.
More informally, the worker will not query its gradient oracle with a nonzero value for some weight unless that weight was already nonzero in the model state of the worker or one of its neighbors at some point in the past.
Similarly, for any $k \in [d]$, if $(\*e_k^\top\*x_{t+1,i}) \ne 0$, then either there exists an $s \le t$ and $j$ such that $(i,j) \in \{(i,i)\} \cup G$ and $(\*e_k^\top\*x_{s,j}) \ne 0$, or one of the gradient oracle's outputs $\*v$ on worker $i$ at iteration $t$ has $\*e_k^\top\*v \ne 0$.
Informally, a worker's model will not have a nonzero weight unless either (1) that weight was nonzero on that worker or one of its neighbors at a previous iteration, or (2) the corresponding entry in one of the gradients the worker sampled at that iteration was nonzero.

Intuitively, we are requiring that algorithm $A$ will not modify those coordinates that remain zero in all previous oracle outputs and neighboring models.\footnote{On the other hand, it is possible to even drop the zero-respecting requirement and extend $A$ to all the deterministic (not in the sense of sampling but the actual executions) algorithms. At a cost, we would need the function class to follow an ``orthogonal invariant'' property, and the model dimension needs to be large enough. We leave this discussion to the appendix.} This lets $A$ use a wide space of accessible information in communication and allows our class to cover first-order methods including SGD~\cite{ghadimi2013stochastic}, Momentum SGD~\cite{nesterov1983method}, Adam~\cite{kingma2014adam}, RMSProp~\cite{tieleman2012lecture}, Adagrad~\cite{ward2018adagrad}, and AdaDelta~\cite{zeiler2012adadelta}.
We let \highlight{algorithm class $\mathcal{A}_{B}$} denote the set of all algorithms $A$ that satisfy these assumptions.


So far our assumptions in this layer cover both centralized and decentralized protocols. Decentralized protocols, however, must satisfy the additional assumption that they communicate via \emph{gossip} (see Section~\ref{section related work}) \cite{boyd2005gossip,boyd2006randomized}.
A single step of gossip protocol can be expressed as
\begin{equation}\label{equation gossip update}
    \textstyle
    \*z_{t,i}\leftarrow\sum_{j\in\mathcal{N}_i}\*y_{t,j}\*W_{ji}, \; \forall i\in[n]
\end{equation}
for some constant doubly stochastic matrix $\*W\in\mathbb{R}^{n\times n}$ called the \emph{communication matrix} and $\* y$ and $\* z$ are the input and output of the gossip communication step, respectively. The essence of a single Gossip step is to take weighted average over the neighborhood specified by a fixed matrix.
To simplify later discussion, we further define the \highlight{gossip matrix class $\mathcal{W}_{n}$} as the set of all matrices $\*W\in\mathbb{R}^{n\times n}$, where $\*W$ is doubly stochastic and $\*W_{ij} \ne 0$ only if $(i,j) \in G$. We call every $\*W\in\mathcal{W}_{n}$ a gossip matrix and we use $\lambda=\max\{|\lambda_2|,|\lambda_n|\} \in [0,1)$ to denote its general second-largest eigenvalue, where $\lambda_i$ denotes the $i$-th largest eigenvalue of $\*W$.
We let \highlight{gossip algorithm class $\mathcal{A}_{B,\*W}$} denote the set of all algorithms $A \in \mathcal{A}_{B}$ that only communicate via gossip using a single matrix $\*W \in \mathcal{W}_{n}$. It trivially holds that $\mathcal{A}_{B,\*W}\subset\mathcal{A}_{B}$.

\subsection{Topology Layer}

Topology-layer assumptions comprise constraints on how workers are connected topologically.
We let the \highlight{graph class $\mathcal{G}_{n,D}$} denote the class of graphs $G$ connecting $n$ workers (vertices) with diameter $D$, where diameter of a graph measures the maximum distance between two arbitrary vertices (so $1 \le D \le n-1$).
A centralized analysis here typically will also require that $G$ be either complete or complete-bipartite (with parameter servers and workers as the two parts): lacking this requirement and allowing arbitrary graphs is what makes an analysis decentralized in the topology layer.


\subsection{Complexity Measures}
Now that we have defined the classes we are interested in, we can use them to define the complexity measures we will bound in our theoretical results.
Given a loss function $f\in\mathcal{F}_{\Delta, L}$, a set of underlying oracles $O\in\mathcal{O}_{\sigma^2}$, a graph $G\in\mathcal{G}_{n,D}$, and an algorithm $A \in \mathcal{A}_B$, let $\*{\hat{x}}_{t}^{A,f,O,G}$ denote the output of algorithm $A$ at the end of iteration $t$ under this setting. 
Then the \emph{iteration complexity} of $A$ solving $f$ under $O$ and $G$ is defined as
\[
    T_\epsilon(A, f, O, G) = \min\left\{t\in\mathbb{N} \,\middle|\, \mathbb{E}\left\|\nabla f(\*{\hat{x}}_{t}^{A,f,O,G})\right\|\leq \epsilon\right\},
\]
that is, the least number of iterations required by $A$ to find a $\epsilon$-stationary-in-expectation point of $f$. 

\section{Lower Bound}\label{section limit}
Given the setup in Section~\ref{section setting}, we can now present and discuss our lower bound on the iteration complexity. 
Note that in the formulation of protocol layer, the algorithm class $\mathcal{A}_B$ only specifies the information available for each worker, and thus $\mathcal{A}_B$ covers both centralization and decentralization in the protocol layer. Here, we show our lower bound in two parts: first a general bound where an arbitrary protocol that follows $\mathcal{A}_B$ is allowed, and then a corollary bound for the case where only decentralized protocol is allowed.
\subsection{Lower Bound for Arbitrary Protocol}
We start from the general bound.
We expect this lower bound to show given arbitrary setting (functions, oracles and graph), the smallest iteration complexity we could obtain from $\mathcal{A}_B$, i.e.
\begin{align}\label{defineequation_lowerbound}
    \adjustlimits\inf_{A\in\mathcal{A}_B}\sup_{f\in\mathcal{F}_{\Delta, L}}\sup_{O\in\mathcal{O}_{\sigma^2}}\sup_{G\in\mathcal{G}_{n,D}}T_\epsilon(A, f, O, G),
\end{align}
it suffices to construct a hard instance containing a loss function $\hat{f}\in\mathcal{F}_{\Delta, L}$, a graph $\hat{G}\in\mathcal{G}_{n,D}$ and a set of oracles $\hat{O}\in\mathcal{O}_{\sigma^2}$ and obtain a valid lower bound on $\inf_{A\in\mathcal{A}_B}T_\epsilon(A, \hat{f}, \hat{O}, \hat{G})$ since Equation~(\ref{defineequation_lowerbound}) is always lower bounded by $\inf_{A\in\mathcal{A}_B}T_\epsilon(A, \hat{f}, \hat{O}, \hat{G})$.

For the construction, we follow the idea of probabilistic zero-chain model \cite{carmon2017lower,carmon2019lower,arjevani2019lower,zhou2019lower}, which is a special loss function where adjacent coordinates are closely dependent on each other like a ``chain.'' Our main idea is to use this function as $f$ and split this chain onto different workers. Then the workers must conduct a sufficient number of optimization steps and rounds of communication to make progress.\footnote{For brevity, we leave details in the supplementary material.} From this, we obtain the following lower bound.
\begin{theorem}\label{thmlowerbound}
For function class $\mathcal{F}_{\Delta, L}$, oracle class $\mathcal{O}_{\sigma^2}$ and graph class $\mathcal{G}_{n,D}$ defined with any $\Delta > 0$, $L > 0$, $n \in \mathbb{N}^{+}$, $D\in \{1,2,\ldots, n-1\}$, $\sigma > 0$, and $B\in\mathbb{N}^{+}$,
there exists $f\in\mathcal{F}_{\Delta, L}$,  $O\in\mathcal{O}_{\sigma^2}$, and $G\in\mathcal{G}_{n,D}$, such that no matter what $A\in\mathcal{A}_B$ is used, $T_\epsilon(A, f, O, G)$ will always be lower bounded by
\begin{equation}
    \label{eqnlowerbound}
    \Omega\left(\frac{\Delta L\sigma^2}{nB\epsilon^4} + \frac{\Delta LD}{\epsilon^2}\right).
\end{equation}
\end{theorem}

\textbf{Dependency on the parameters.} The bound in Theorem~\ref{thmlowerbound} consists of a sample complexity term, which is the dominant one for small $\epsilon$, and a communication complexity term. We can see the increase of query budget $B$ will only reduce the sample complexity. On the other hand, as the diameter $D$ of a graph will generally increase as the number of vertices $n$ increases, we can observe a trade-off between two terms when the system scales up: when more workers join the system, the communication complexity will gradually become the dominant term.

\textbf{Consistency with the literature.} Theorem~\ref{thmlowerbound} is tightly aligned with the state-of-the-art bounds in many settings. With $n=B=D=1$, we recover the tight bound for sequential stochastic non-convex optimization $\Theta(\Delta L\sigma^2\epsilon^{-4})$ as shown in \citet{arjevani2019lower}. With $\sigma=0, D=1$, we recover the tight bound for sequential non-stochastic non-convex optimization $\Theta(\Delta L\epsilon^{-2})$ as shown in \citet{carmon2019lower}. 
With $B=1, D=1$, we recover the tight bound for centralized training $\Theta(\Delta L\sigma^2(n\epsilon^4)^{-1})$ given in \citet{li2014communication}.

\textbf{Improvement upon previous results.}
Previous works like
\citet{seaman2017optimal,scaman2018optimal} provide similar lower bounds in a convex setting which relates to the diameter. However, these results treat $D$ as a fixed value, i.e., $D=n-1$, and thus makes the bound to be only tight on linear graph. By comparison, Theorem~\ref{thmlowerbound} allows $D$ to be chosen independently to $n$.
\subsection{Lower Bound for Decentralized Protocol}
The bound in Theorem~\ref{thmlowerbound} holds for both centralized and decentralized protocols.
A natural question is: \textit{How would the lower bound adapt if the protocol is restricted to be decentralized?} i.e., the quantity of
\begin{align*}
    \adjustlimits\inf_{A\in\text{\highlight{$\mathcal{A}_{B,\*W}$}}}\sup_{f\in\mathcal{F}_{\Delta, L}}\sup_{O\in\mathcal{O}_{\sigma^2}}\sup_{G\in\mathcal{G}_{n,D}}T_\epsilon(A, f, O, G),
\end{align*}
we can extend the lower bound to Gossip in the following corollary.
\begin{corollary}\label{corollary lower bound}
    For every $\Delta > 0$, $L > 0$, $n \in \{2, 3, 4, \cdots\}$, $\sigma > 0$, and $B\in\mathbb{N}^{+}$,
    there exists a loss function $f\in\mathcal{F}_{\Delta, L}$, a set of underlying oracles $O\in\mathcal{O}_{\sigma^2}$, a gossip matrix $\*W\in\mathcal{W}_n$ with second largest eigenvalue being $\lambda=\cos(\pi/n)$, and a graph $G\in\mathcal{G}_{n,D}$, such that no matter what $A\in\mathcal{A}_{B,\*W}$ is used, $T_\epsilon(A, f, O, G)$ will always be lower bounded by
    \begin{equation}
    \Omega\left(\frac{\Delta L\sigma^2}{nB\epsilon^4} + \frac{\Delta L}{\epsilon^2\sqrt{1-\lambda}}\right).
    \end{equation}
\end{corollary}
\textbf{Gap in the existing algorithms.}
Comparing this lower bound with many state-of-the-art decentralized algorithms (Table~\ref{table complexity}), we can see they match on the sample complexity but leave a gap on the communication complexity. In many cases, the spectral gap significantly depends on the number of workers $n$ and thus can be arbitrarily large. For example, when the graph $G$ is a cycle graph or a linear graph, the gap of those baselines can increase by up to $O(n^6)$ \cite{brooks2011handbook,gerencser2011markov}!
\section{DeFacto: Optimal Complexity in Theory}\label{section optimal}
\begin{algorithm}[t]
\small
\begingroup
    \setlength{\abovedisplayskip}{2pt}
    \setlength{\belowdisplayskip}{2pt}
	\caption{Decentralized Stochastic Gradient Descent with Factorized Consensus Matrices (DeFacto) on worker $i$}\label{DeFacto}
	\begin{algorithmic}[1]
		\Require initialized model $\*x_{0,i}$, a copy of model $\tilde{\*x}_{0,i}\leftarrow\*x_{0,i}$, gradient buffer $\*g=\*0$, step size $\alpha$, a sequence of communication matrices $\{\*W_r\}_{1\leq r\leq R}$ of size $R$, number of iterations $T$, neighbor list $\mathcal{N}_i$
		
		\For{$t=0,1,\cdots,T-1$}
		    \State $k\leftarrow \lfloor t/2R\rfloor$.
		    \State $r\leftarrow t\bmod 2R$.
		    \If{$0\leq r<R$}
		        \State Spend all $B$ oracle budgets to compute stochastic gradient $\tilde{\*g}$ at point $\*x_{k,i}$ and accumulate it to gradient buffer: $\*g\leftarrow\*g+\tilde{\*g}$.
		    \Else
		        \State Update model copy with the $r$-th matrix in $\{\*W_r\}_{1\leq r\leq R}$:
    			\begin{align}
    			\tilde{\*x}_{t+1,i}\leftarrow\sum_{j\in\mathcal{N}_i\cup\{i\}}\tilde{\*x}_{t,j}[\*W_r]_{ji}
    			\end{align}
		    \EndIf
		    \If{$r=2R-1$}
		        \State Update Model: $\*x_{t+1,i}\leftarrow \tilde{\*x}_{t+1,i}-\alpha\frac{\*g}{R}$.
		        \State Reinitialize gradient buffer: $\*g\leftarrow\*0$.
		        \State Copy the current model: $\tilde{\*x}_{t+1,i}\leftarrow\*x_{t+1,i}$.
		    \EndIf
		\EndFor
		\State \textbf{return} $\*{\hat{x}}=\frac{1}{n}\sum_{i=1}^{n}\*x_{T,i}$
	\end{algorithmic}
	\endgroup
\end{algorithm}
\begin{algorithm}[t]
\small
\begingroup
    \setlength{\abovedisplayskip}{2pt}
    \setlength{\belowdisplayskip}{2pt}
	\caption{Decentralized Stochastic Gradient Tracking with By-Phase Accelerated Gossip (DeTAG) on worker $i$}\label{DeTAG}
	\begin{algorithmic}[1]
		\Require initialized model $\*x_{0,i}$, a copy of model $\tilde{\*x}_{0,i}\leftarrow\*x_{0,i}$, gradient tracker $\*y_{0,i}$, gradient buffer $\*g_{(0)}=\*g_{(-1)}=\*0$, step size $\alpha$, a gossip matrix $\*W$, number of iterations $T$, neighbor list $\mathcal{N}_i$
		
		\For{$t=0,1,\cdots,T-1$}
		    \State $k\leftarrow \lfloor t/R\rfloor$.
		    \State $r\leftarrow t\bmod R$.
		    \State Perform the $r$-th step in Accelerate Gossip:
		        \begin{align}
		            & \tilde{\*x}_{t+1,i} \leftarrow AG(\tilde{\*x}_{t,i}, \*W, \mathcal{N}_i, i)\\
		            & \*y_{t+1,i} \leftarrow AG(\*y_{t,i}, \*W, \mathcal{N}_i, i)
		        \end{align}
		    \State Spend all $B$ oracle budgets to compute stochastic gradient $\tilde{\*g}$ at point $\*x_{k,i}$ and accumulate it to gradient buffer: $\*g_{(k)}\leftarrow\*g_{(k)}+\tilde{\*g}$.
		    \If{$r=R-1$}
		        \State Update gradient tracker and model:
		            \begin{align}
		                & \*x_{t+1,i}\leftarrow \tilde{\*x}_{t+1,i} - \alpha\*y_i\\
		                & \*y_{t+1,i}\leftarrow \*y_{t+1,i} + \*g_{(k)} - \*g_{(k-1)}
		            \end{align}
		        \State Reinitialize gradient buffer: $\*g_{(k-1)}\leftarrow\*g_{(k)}$ and then $\*g_{(k)}\leftarrow\*0$.
		        \State Copy the current model: $\tilde{\*x}_{t+1,i}\leftarrow\*x_{t+1,i}$.
		    \EndIf
		\EndFor
		\State \textbf{return} $\*{\hat{x}}=\frac{1}{n}\sum_{i=1}^{n}\*x_{T,i}$
	\end{algorithmic}
	\endgroup
\end{algorithm}
\begin{algorithm}[t]
	\caption{Accelerated Gossip (AG) with R steps}\label{AG}
	\begin{algorithmic}[1]
		\Require $\*z_{0,i}$, $\*W$, $\mathcal{N}_i$, $i$
		\State $\*z_{-1,i}\leftarrow\*z_{0,i}$
		\State $\eta\leftarrow\frac{1-\sqrt{1-\lambda^2}}{1+\sqrt{1-\lambda^2}}$
		\For{$r=0,1,2,\cdots,R-1$}
		    \State $\*z_{r+1,i} \leftarrow (1+\eta)\sum_{j\in\mathcal{N}_i\cup\{i\}}\*z_{r,j}\*W_{ji} - \eta\*z_{r-1,i}$
		\EndFor
		\State \textbf{return} $\*z_{R,i}$
	\end{algorithmic}
\end{algorithm}
In the previous section we show the existing algorithms have a gap compared to the lower bound. This gap could indicate the algorithms are suboptimal, but it could also be explained by our lower bound being loose.
In this section we address this issue by proposing DeFacto, an example algorithm showing the lower bound is achievable, which verifies the tightness of our lower bound---showing that (\ref{eqnlowerbound}) would hold with equality and $\Theta(\cdot)$, not just $\Omega(\cdot)$. 

We start with the following insight on the theoretical gap: the goal of communication is to let all the workers obtain information from neighbors.
Ideally, the workers would, at each iteration, perform (\ref{equation gossip update}) with $\*W^*=\*1_n\*1_n^\top/n$, where $\*1_n$ is the $n$-dimensional all-one vector.
We call this matrix the \emph{Average Consensus} matrix. The Average Consensus is statistically equivalent to centralized communication (All-Reduce operation).
However, due to the graph constraints, we can not use this $\*W^*$ unless workers are fully connected; instead, a general method is to repeatedly apply a sequence communication matrices in consecutive iterations and let workers achieve or approach the Average Consensus. Previous work uses Gossip matrix $\*W$ and expect $\prod_{r=1}^{R}\*W \approx \*1_n\*1_n^\top/n$ for some $R$. This $R$ is known to be proportional to the mixing time of the Markov Chain $\*W$ defines~\cite{lu2020mixml,lu2020moniqua}, which is related to the inverse of its spectral gap~\cite{levin2017markov}.
This limits convergence depending on the spectrum of the $\*W$ chosen.
The natural question to ask here is: can we do better? What are the limits of how fast we can reach average consensus on a connectivity graph $G$?
This question is answered by the following lemma.
\begin{lemma}\label{lemma_definitive_consensus}
For any $G\in\mathcal{G}_{n, D}$, let $\mathcal{W}_{G}$ denote the set of $n\times n$ matrices such that for all $\*W\in\mathcal{W}_{G}$, $\*W_{ij}=0$ if edge $(i,j)$ does not appear in $G$. There exists a sequence of $R$ matrices $\{\*W_r\}_{r\in[R]}$ that belongs to $\mathcal{W}_{G}$ such that $R\in\{D, D+1, \cdots, 2D\}$ and
\[
    \*W_{R-1}\*W_{R-2}\cdots \*W_0 = \frac{\*1_n\*1_n^\top}{n} = \* W^*.
\]
\end{lemma}
Lemma~\ref{lemma_definitive_consensus} is a classic result in the literature of graph theory. The formal proof and detailed methods to identify these matrices can be found in many previous works \cite{georgopoulos2011definitive,ko2010matrix,hendrickx2014graph}. 
Here we treat this as a black box procedure.\footnote{We cover specific algorithms and details in the supplementary.}

Lemma~\ref{lemma_definitive_consensus} shows that we can achieve the exact average consensus by factorizing the matrix $\*1_n\*1_n^\top/n$, and we can obtain the factors from a preprocessing step. From here, the path to obtain an optimal rate becomes clear: starting from $t=0$, workers first spend $R$ iterations only computing stochastic gradients and then another $R$ iterations to reach consensus communicating via factors from Lemma~\ref{lemma_definitive_consensus}; they then repeat this process until a stationary point is found. We call this algorithm DeFacto (Algorithm~\ref{DeFacto}).

DeFacto is statistically equivalent to centralized SGD operating $T/2R$ iterations with a mini-batch size of $BR$.
It can be easily verified that DeFacto holds membership in $\mathcal{A}_B$. A straightforward analysis gives
the convergence rate of DeFacto shown in the following Theorem.
\begin{theorem}\label{thmrateDeFacto}
    Let $A_1$ denote Algorithm~\ref{DeFacto}. For $\mathcal{F}_{\Delta, L}$, $\mathcal{O}_{\sigma^2}$ and $\mathcal{G}_{n,D}$ defined with any $\Delta > 0$, $L > 0$, $n \in \mathbb{N}^{+}$, $D\in \{1,2,\ldots,n-1\}$, $\sigma > 0$, and $B\in\mathbb{N}^{+}$, the convergence rate of $A_1$ running on any loss function $f\in\mathcal{F}_{\Delta,L}$, any graph $G\in\mathcal{G}_{n,D}$, and any oracles $O\in\mathcal{O}_{\sigma^2}$ is bounded by
    \begin{equation}
        T_\epsilon(A_1, f, O, G) \leq O\left(\frac{\Delta L\sigma^2}{nB\epsilon^4} + \frac{\Delta LD}{\epsilon^2}\right).
    \end{equation}
\end{theorem}
Comparing Theorem~\ref{thmlowerbound} and Theorem~\ref{thmrateDeFacto}, DeFacto achieves the optimal rate asymptotically. \emph{This shows that our lower bound in Theorem~\ref{thmlowerbound} is tight.}

Despite its optimality, the design of DeFacto is unsatisfying in three aspects: (1) It compromises the throughput\footnote{The number of stochastic gradients computed per iteration.} by a factor of two because in each iteration, a worker either communicates with neighbors or computes gradients but not both. This fails to overlap communication and computation and creates extra idle time for the workers. (2) It needs to iterate over all the factor matrices before it can query the gradient oracle at subsequent parameters. When diameter $D$ increases, the total time to finish such round will increase proportionally.
(3) DeFacto works with decentralized data and arbitrary graph, achieving decentralization in both application and topology layers. However, the matrices used in Lemma~\ref{lemma_definitive_consensus} are not Gossip matrices as defined in $\mathcal{W}_n$, and thus it fails to be decentralized in the protocol-layer sense.

\section{DeTAG: Optimal Complexity in Practice}\label{section detag}
To address the limitations of DeFacto,
a natural idea is to replace all the factor matrices in Lemma~\ref{lemma_definitive_consensus} with a gossip matrix $\*W$. 
The new algorithm after this mild modification is statistically equivalent to a D-PSGD variant: every $R$ iterations, it updates the model the same as one iteration in D-PSGD with a mini-batch size of $BR$ and communicate with a matrix $\*W'$ whose second largest eigenvalue $\lambda'=\lambda^R$, with $T/R$ iterations in total.
However, even with arbitrarily large $R$, the communication complexity in this ``updated D-PSGD'' is still $O(\Delta Ln\varsigma\epsilon^{-2})$ (Table~\ref{table complexity}), leaving an $O(n\varsigma)$ gap compared to our lower bound. 

To close this gap, we adopt two additional techniques:\footnote{Note that neither of these techniques is our original design, and we do not take credit for them. Our main contribution here is to prove their combination leads to optimal complexity.} one is a gradient tracker $\*y$ that is used as reference capturing gradient difference in the neighborhood; the other is using acceleration in gossip as specified in Algorithm~\ref{AG}. Modifying DeFacto results in Algorithm~\ref{DeTAG}, which we call DeTAG. DeTAG works as follows: it divides the total number of iterations $T$ into several phases where each phase contains $R$ iterations. In each iteration, the communication process calls Accelerated Gossip to update a model replica $\tilde{\*x}$ and the gradient tracker (line 4) while the computation process constantly computes gradients at the same point (line 5). At the end of each phase, model $\*x$, its replica $\tilde{\*x}$ and gradient tracker $\*y$ are updated in line 7-10 and then DeTAG steps into the next phase.
Aside from the two additional techniques, the main difference between DeTAG and DeFacto is that the communication matrix in DeTAG is a fixed gossip matrix $\*W$, which allows DeTAG to benefit from decentralization in the protocol layer as well as to adopt arbitrary $R\geq 1$ in practice (allowing $R$ to be tuned independently of $G$).

\textbf{Improvement on design compared to baselines.}
Comparing with other baselines in Table~\ref{table complexity}, the design of DeTAG improves in the sense that (1) It removes the dependency on the outer variance $\varsigma$. (2) It drops the requirement\footnote{\citet{tang2018d} requires the gossip matrix to be symmetric and its smallest eigenvalue is lower bounded by $-\frac{1}{3}$.} on the gossip matrix assumed in \citet{tang2018d}. (3) The baseline DSGT \cite{zhang2019decentralized} and GT-DSGD \cite{xin2021improved} can be seen as special cases of taking $R=1$ and $\eta=0$ in DeTAG. That implies in practice, a well tuned DeTAG can never perform worse than the baseline DSGT or GT-DSGD.

The convergence rate of DeTAG is given in the following theorem.
\begin{theorem}\label{thmrateDeTAG}
    Let $A_2$ denote Algorithm~\ref{DeTAG}. For $\mathcal{F}_{\Delta, L}$, $\mathcal{O}_{\sigma^2}$ and $\mathcal{G}_{n,D}$ defined with any $\Delta > 0$, $L > 0$, $n \in \mathbb{N}^{+}$, $\lambda\in[0,1)$, $\sigma > 0$, and $B\in\mathbb{N}^{+}$, under the assumption of Equation~(\ref{assumeweakclosedistribution}), if we set the phase length $R$ to be
    \[
        R = \frac{\max\left( 
            \frac{1}{2} \log(n),
            \frac{1}{2} \log\left( \frac{\varsigma_0^2 T}{\Delta L} \right)
        \right)}{\sqrt{1-\lambda}},
    \]
    the convergence rate of $A_2$ running on any loss function $f\in\mathcal{F}_{\Delta,L}$, any graph $G\in\mathcal{G}_{n,D}$, and any oracles $O\in\mathcal{O}_{\sigma^2}$ is bounded by
    \[
        T_\epsilon(A_2, f, O, G)\leq O\left(\frac{\Delta L\sigma^2}{nB\epsilon^4} + \frac{\Delta L\log\left(n + \frac{\varsigma_0n}{\epsilon\sqrt{\Delta L}}\right)}{\epsilon^2\sqrt{1-\lambda}} \right).
    \]
\end{theorem}
Comparing Theorem~\ref{thmlowerbound} and
Theorem~\ref{thmrateDeTAG}, DeTAG achieves the optimal complexity with only a logarithm gap.

\textbf{Improvement on complexity.}
Revisiting Table~\ref{table complexity}, we can see the main improvement of DeTAG's complexity is in the two terms on communication complexity: (1) DeTAG only depends on the outer variance term $\varsigma_0$ inside a log, and (2) It reduces the dependency on the spectral gap $1-\lambda$ to the lower bound of square root, as shown in Corollary~\ref{corollary lower bound}.

\textbf{Understanding the phase length $R$.}
In DeTAG, the phase length $R$ is a tunable parameter. Theorem~\ref{thmrateDeTAG} provides a suggested value for $R$. 
Intuitively, the value of $R$ captures the level of consensus of workers should reach before they step into the next phase.
Theoretically, we observe $R$ is closely correlated to the mixing time of $\*W$: if we do not use acceleration in Gossip, then $R$ will become $\tilde{O}\left(\frac{1}{1-\lambda}\right)$, which is exactly the upper bound on the mixing time of the Markov Chain $\*W$ defines \cite{levin2017markov}.

\begin{figure*}[ht!]
  \centering
  \subfigure[100\% Shuffled CIFAR10]{
    \includegraphics[width=0.45\textwidth]{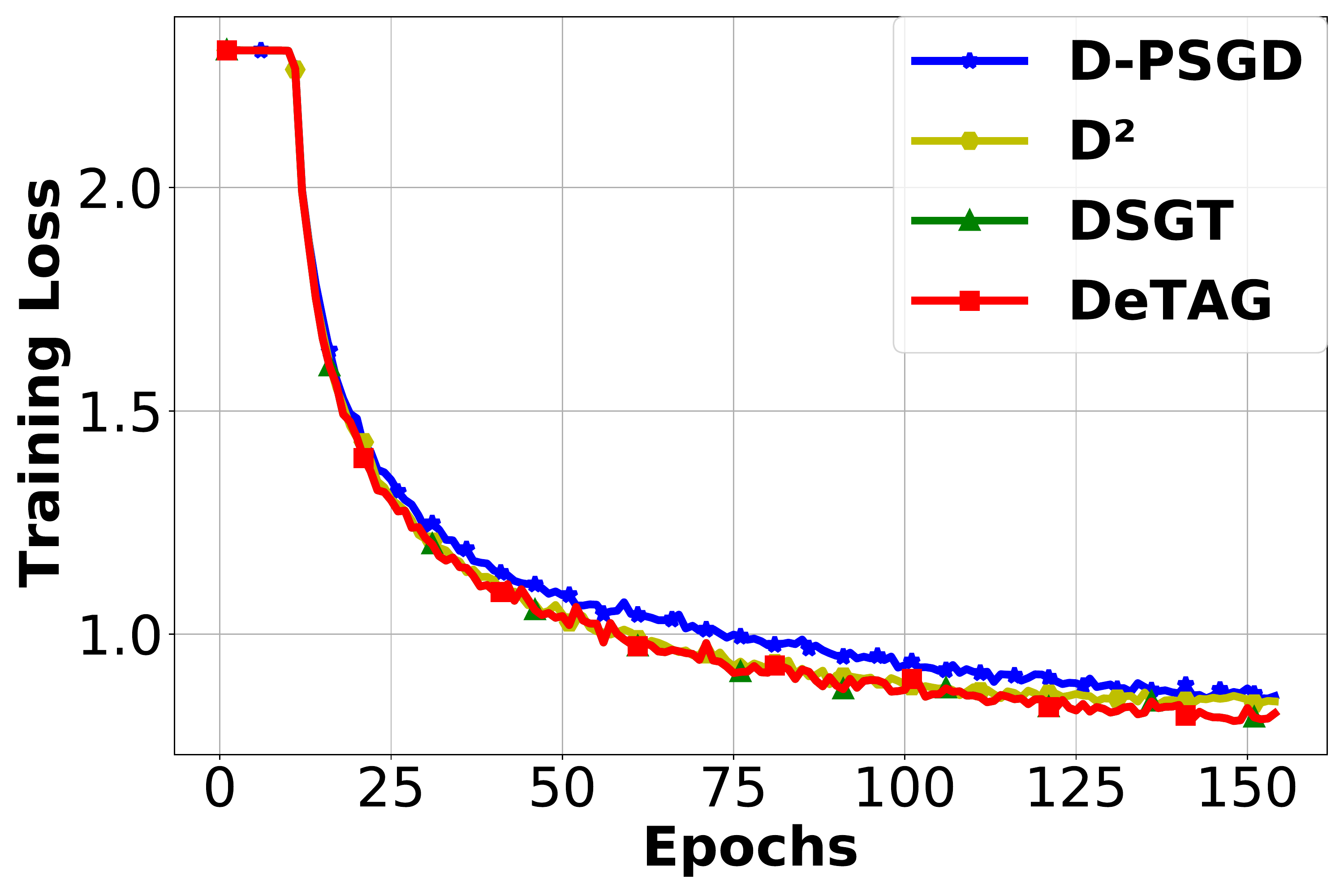}
    \label{exp_lenet_random}
    }
    \subfigure[50\% Shuffled CIFAR10]{
    \includegraphics[width=0.45\textwidth]{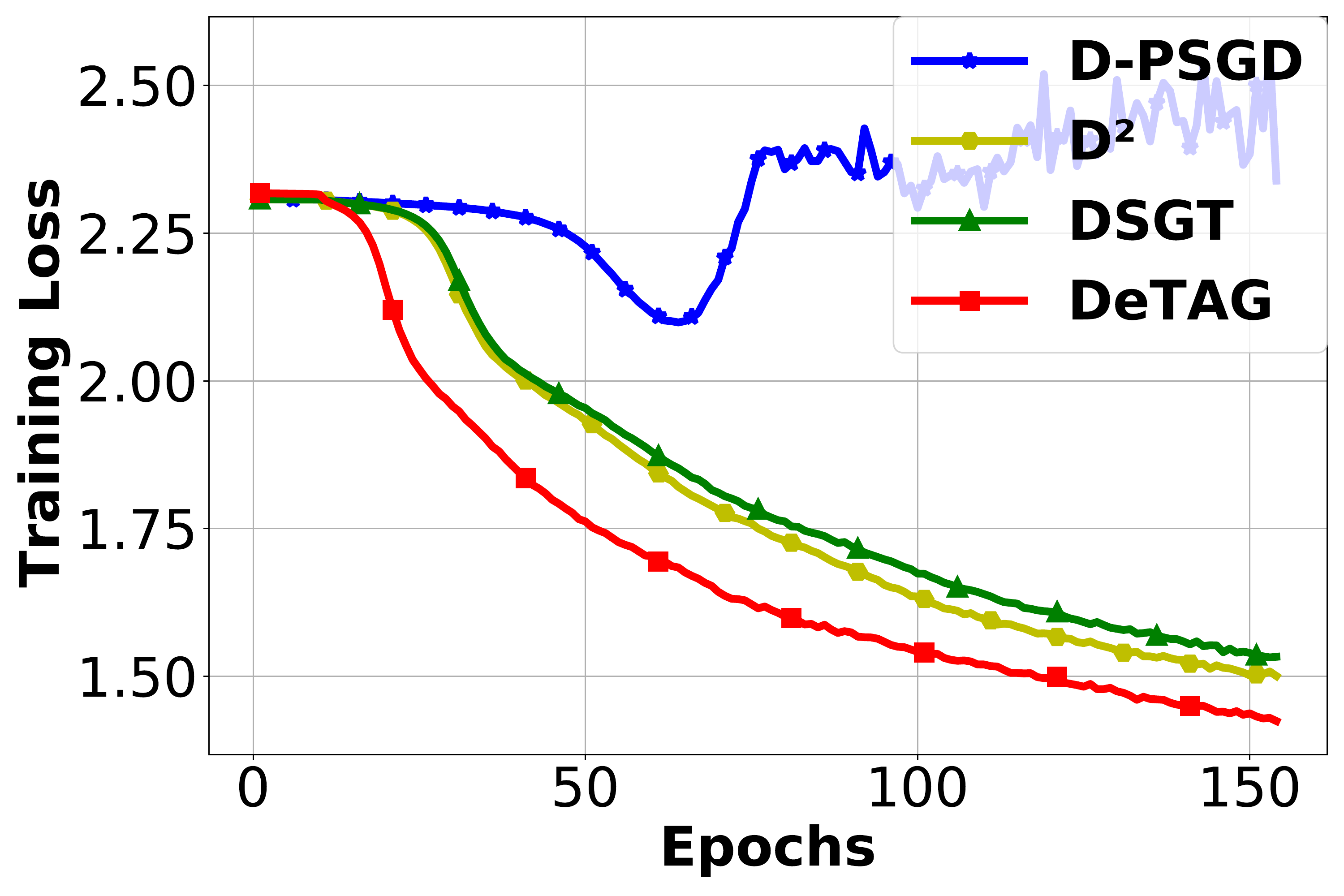}
    \label{exp_lenet_sort50}
  }
  \subfigure[25\% Shuffled CIFAR10]{
    \includegraphics[width=0.45\textwidth]{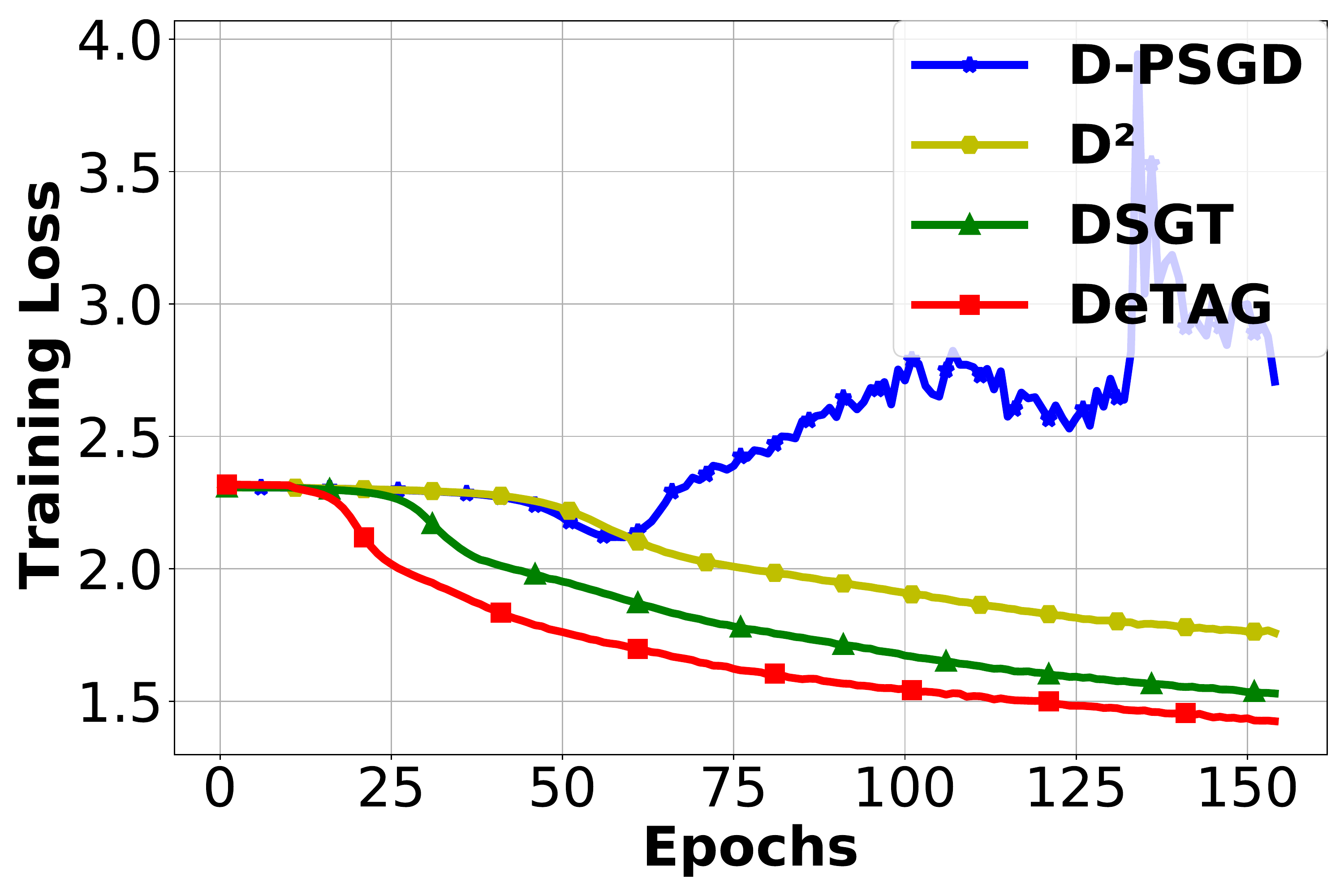}
    \label{exp_lenet_sort75}
  }
  \subfigure[0\% Shuffled CIFAR10]{
    \includegraphics[width=0.45\textwidth]{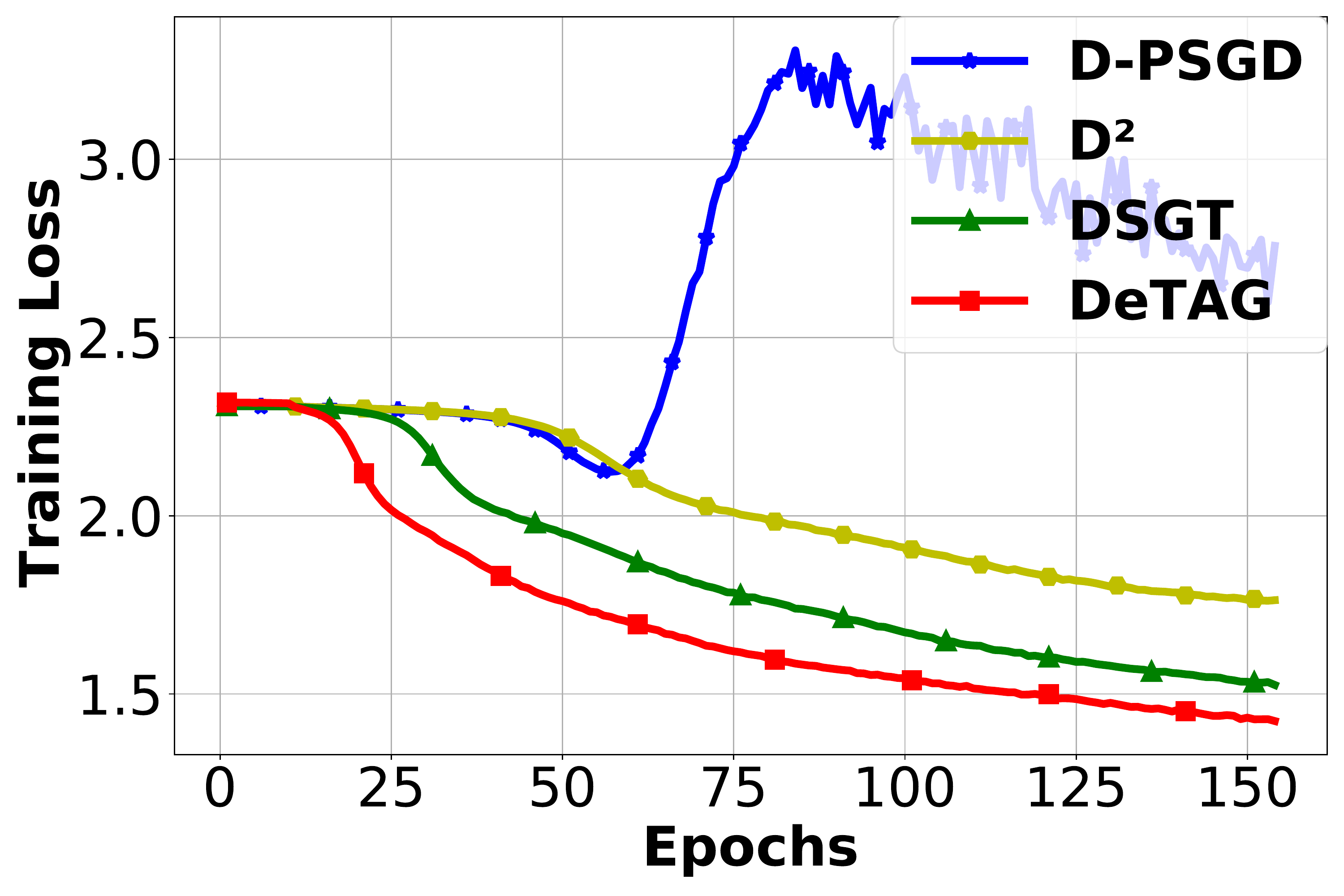}
    \label{exp_lenet_sort100}
    }
  \caption{Fine tuned results of training LeNet on CIFAR10 with different shuffling strategies.}
  \label{lenet_cifar}
\end{figure*}

\begin{figure*}[ht!]
  \centering
  \subfigure[$\kappa=1$ ($1-\lambda\approx$ 4e-2)]{
    \includegraphics[width=0.45\textwidth]{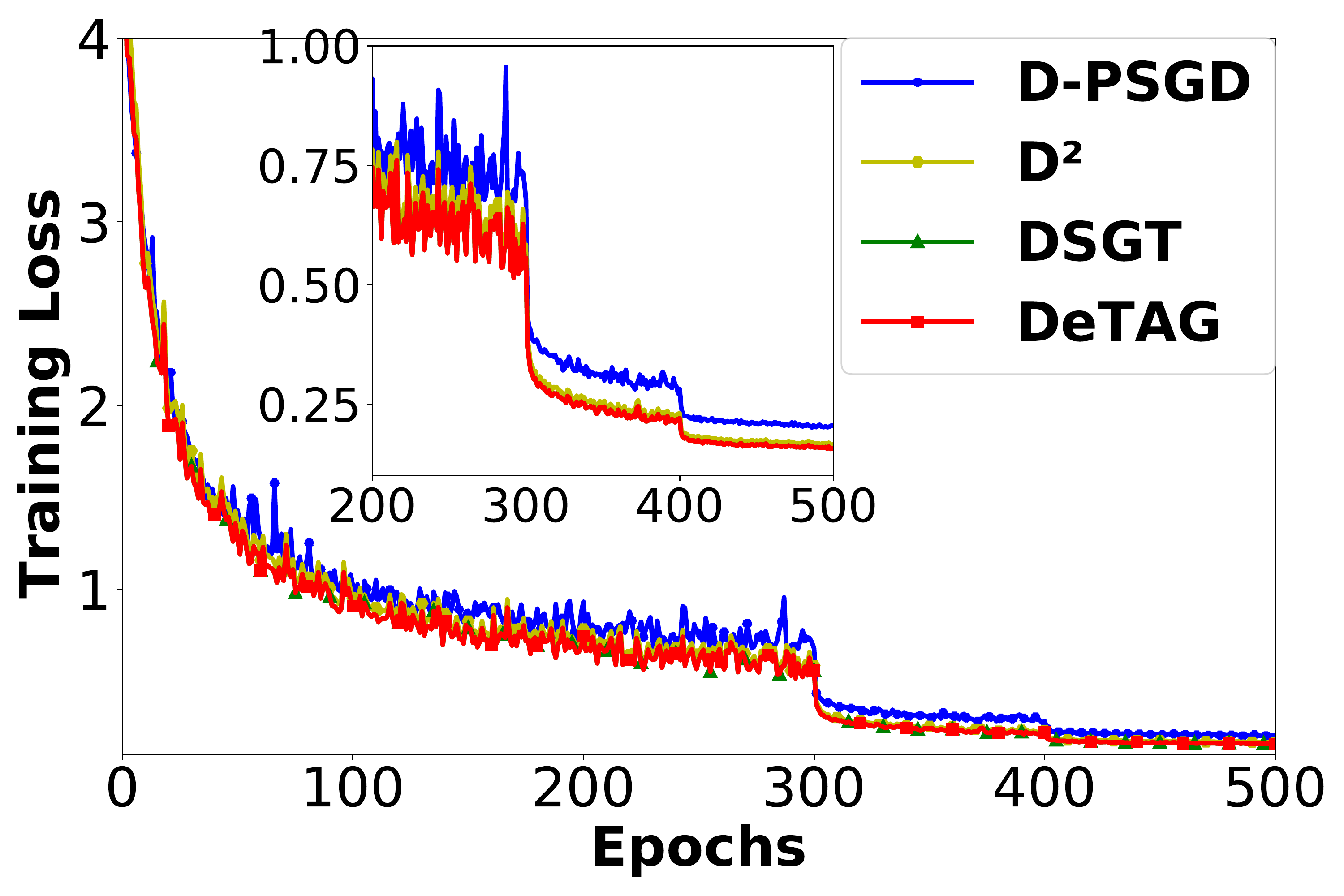}
    }
  \subfigure[$\kappa=0.1$ ($1-\lambda\approx$ 4e-3)]{
    \includegraphics[width=0.45\textwidth]{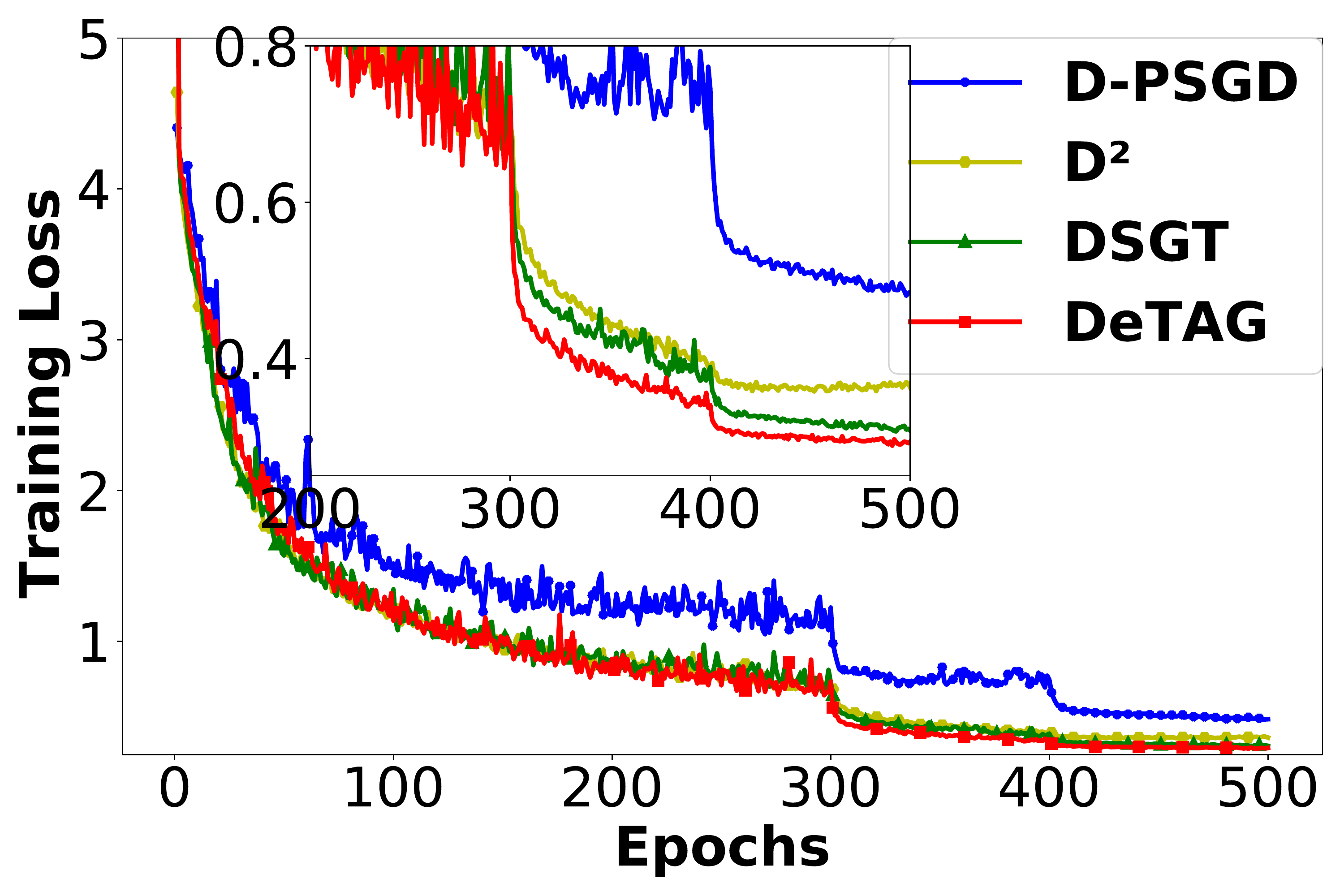}
    }
  \subfigure[$\kappa=0.05$ ($1-\lambda\approx$ 2e-3)]{
    \includegraphics[width=0.45\textwidth]{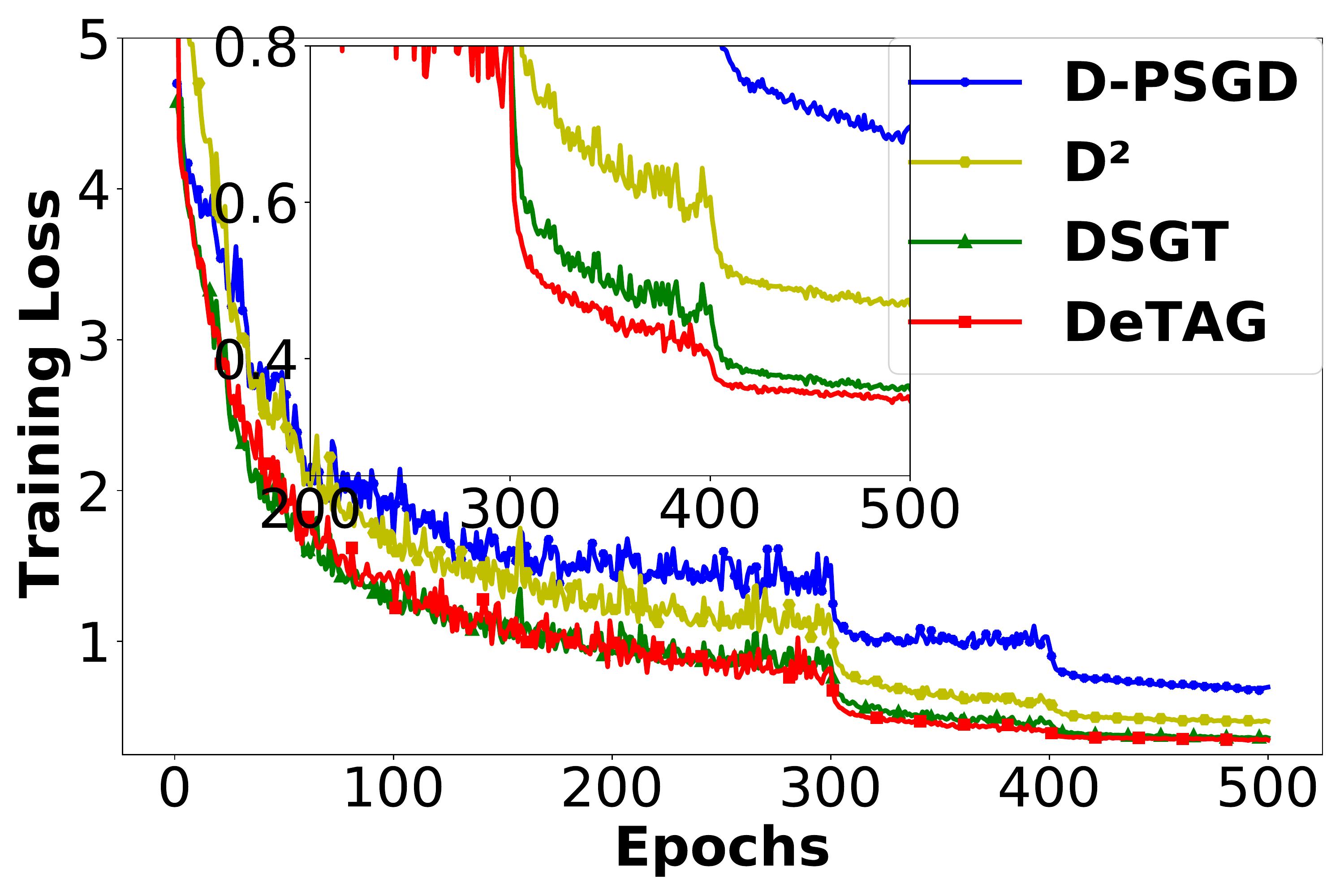}
    }
    \subfigure[$\kappa=0.01$ ($1-\lambda\approx$ 4e-4)]{
    \includegraphics[width=0.45\textwidth]{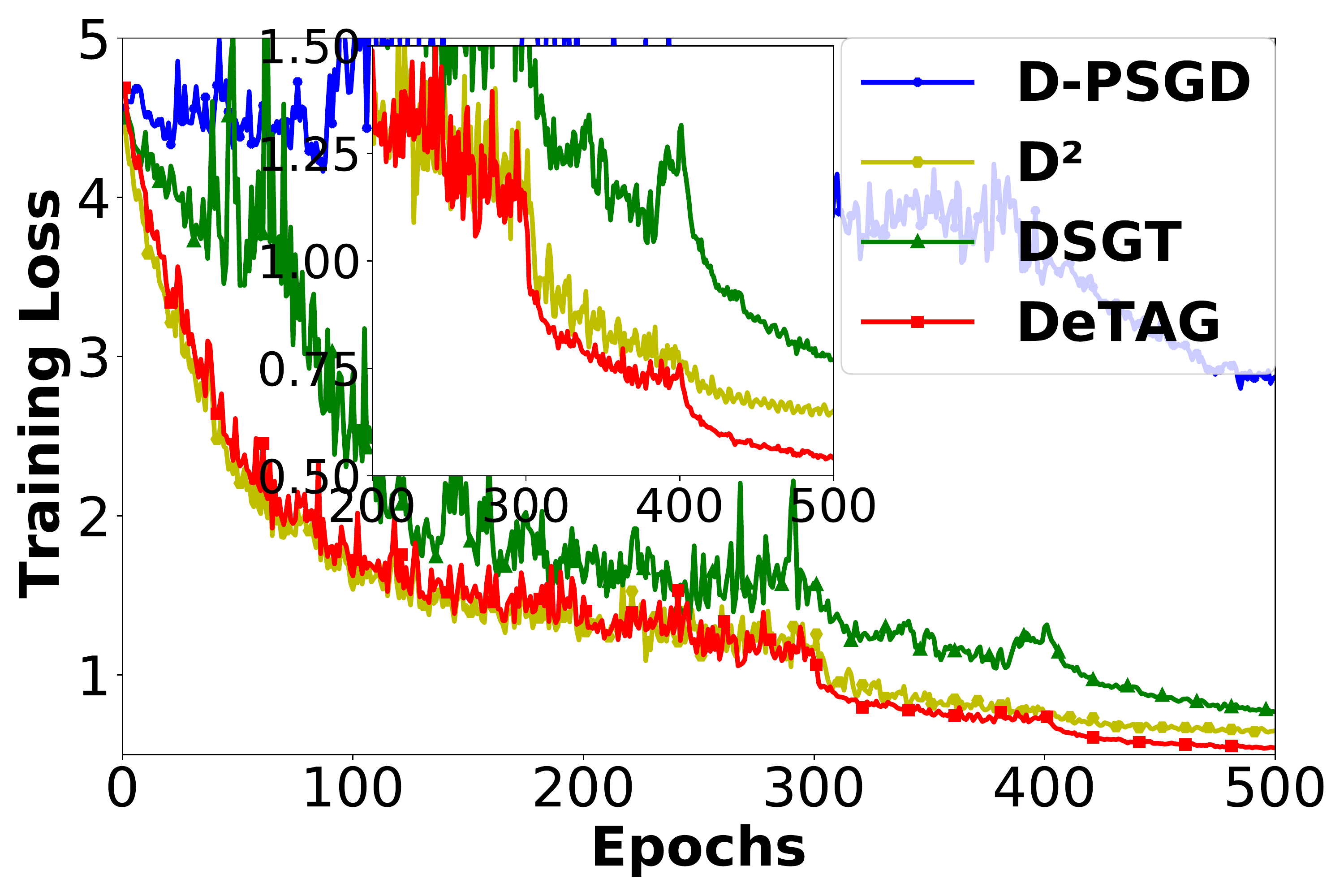}
  }
  \caption{Fine tuned results of training Resnet20 on CIFAR100 with different spectral gaps.}
  \label{resnet20_cifar}
\end{figure*}

\section{Experiments}\label{section experiment}
In this section we empirically compare the performance among different algorithms. 
All the models and training scripts in this section are implemented in PyTorch and run on an Ubuntu 16.04 LTS cluster using a SLURM workload manager running CUDA 9.2, configured with 8 NVIDIA GTX 2080Ti GPUs. We launch one process from the host as one worker and let them use \texttt{gloo} as the communication backend.
In each experiment, we compare the following algorithms\footnote{Since DeFacto is a only a "motivation" algorithm and in practice we observe it performs bad, we do not include the discussion of that.}: D-PSGD \cite{lian2017can}, D$^2$ \cite{tang2018d}, DSGT \cite{zhang2019decentralized} and DeTAG.
Note that GT-DSGD \cite{xin2021improved} and DSGT \cite{zhang2019decentralized} are essentially the same algorithm so we omit the comparison to GT-DSGD.
Also note that SGP \cite{assran2019stochastic} reduces to D-PSGD for symmetric mixing matrices in undirected graphs.
Throughout the experiment we use Ring graph.
Hyperparameters can be found in the supplementary material.
\paragraph{Convergence over different outer variance.}
In the first experiments, we investigate the correlation between convergence speed and the outer variance $\varsigma(\varsigma_0)$. 
We train LeNet on CIFAR10 using 8 workers, which is a standard benchmark experiment in the decentralized data environment \cite{tang2018d,zhang2019decentralized}. To create the decentralized data, we first sort all the data points based on its labels, shuffle the first $X\%$ data points and then evenly split to different workers. The $X$ controls the degree of decentralization, we test $X=0, 25, 50, 100$ and plot the results in Figure~\ref{lenet_cifar}.

We can see in Figure~\ref{exp_lenet_random} when the dataset is fully shuffled, all the algorithms converge at similar speed while D-PSGD converges a little slower than other variance reduced algorithms. From Figure~\ref{exp_lenet_sort50} to Figure~\ref{exp_lenet_sort100} we can see when we shuffle less portion of the dataset, i.e., the dataset becomes more decentralized, D-PSGD fails to converge even with fine-tuned hyperparameter. Meanwhile, among D$^2$, DSGT and DeTAG, we can see DeTAG converges the fastest. When dataset becomes more decentralized, DSGT seems to receive more stable performance than D$^2$.
\paragraph{Convergence over different spectral gaps.}
In the second experiments, we proceed to explore the relation between convergence speed and spectral gap $1-\lambda$ of the gossip matrix $\*W$. We use 16 workers connected with a Ring graph to train Resnet20 on CIFAR100, and we generate a $\*W_0$ on such graph using Metropolis method. Then we adopt the slack matrix method to modify the spectral gap \cite{lu2020mixml}: $\*W_{\kappa}=\kappa\*W_0 + (1-\kappa)\*I$, where $\kappa$ is a control parameter.
We test $\kappa=1, 0.1, 0.05, 0.01$ and plot the results in Figure~\ref{resnet20_cifar}.
We can see with different $\kappa$, DeTAG is able to achieve faster convergence compared to baselines. When the network becomes sparse, i.e., $\kappa$ decreases, DeTAG enjoys more robust convergence.

\section{Conclusion}\label{section conclusions}
In this paper, we investigate the tight lower bound on the iteration complexity of decentralized training. We propose two algorithms, DeFacto and DeTAG, that achieve the lower bound in terms of different decentralization in a learning system. DeTAG uses Gossip protocol, and is shown to be empirically competitive to many baseline algorithms, such as D-PSGD.
In the future, we plan to investigate the variants of the complexity bound with respect to communication that are compressed, asynchronous, etc.

\section*{Acknowledgement}
This work is supported by NSF IIS-2046760. The authors would like to thank A. Feder Cooper, Jerry Chee, Zheng Li, Ran Xin, Jiaqi Zhang and anonymous reviewers from ICML 2021 for providing valuable feedbacks on earlier versions of this paper.

\bibliography{main}

\newpage
\begin{center}
{\huge\textbf{Supplementary Material}}
\end{center}

\appendix
\section{Experimental Details}
\subsection{Hyperparameter Tuning}
In the experiment of training LeNet on CIFAR10, we tune the step size using grid search inside the following range: \{5e-3, 1e-3, 5e-4, 2.5e-4, 1e-4, 5e-5\}. Note that this range is in general smaller than the one chosen in \cite{zhang2019decentralized}, since here we are working with unshuffled data, and we found original range in baselines causes algorithms to diverge easily. Following \cite{tang2018d}, we let each run warm up for 10 epochs with step size 1e-5.
For DeTAG, we further tune the accelerated gossip parameter $\eta$ within \{0, 0.1, 0.2, 0.4\} and phase length $R$ within \{1, 2, 3\}. We fix the momentum term to be 0.9 and weight decay to be 1e-4.

In the experiment of training Resnet20 on CIFAR100, we tune the step size using grid search inside the following range: \{0.5, 0.1, 0.05, 0.01, 0.005\}.
For DeTAG, we further tune the accelerated gossip parameter $\eta$ within \{0, 0.1, 0.2, 0.4\} and phase length $R$ within \{1, 2, 3\}. We fix the momentum term to be 0.9 and weight decay to be 5e-4.

The hyperparameters adopted for each runs are shown in Table~\ref{table:hyperparameter:stepsize} and Table~\ref{table:hyperparameter:detag}.

\subsection{Techniques of Running DeTAG}
We can see in the main loop of DeTAG, several gradient queries are made at the same point. This essentially is equivalent to a large mini-batch size. In practice, however, we can modify this to use local-steps and get better empirical results \cite{lin2018don}. Another technique is to use warm-up epochs when data is decentralized. We observe it ensures a smooth convergence in practice.
Last but not least, since at first the noise in the algorithms is generally large, we can use a dynamic phase length to obtain better results. That is, we start from phase length 1 for the first few epochs, and let DeTAG follow the special case of DSGT. Then we can gradually increase the phase length following given policies. The intuition is that as algorithm converges, we would need less noise from communication, and thus a longer phase length can benefit.

\begin{table}[ht]
\small
\caption{(Initial) Step size $\alpha$ used for each experiments.}
\label{table:hyperparameter:stepsize}
\small
\begin{center}
\begin{tabular}{ccccccc}
\toprule
\multirow{2}{*}{Experiment} & \multirow{2}{*}{Setting} & \multicolumn{4}{c}{Algorithm} \\
\cmidrule{3-6}
& & D-PSGD & D$^2$ & DSGT & DeTAG\\
\midrule
\multirow{4}{*}{LeNet/CIFAR10} & 100\% Shuffled & 5e-3 & 5e-3 & 5e-3 & 5e-3 \\
& 50\% Shuffled & 5e-5 & 2.5e-4 & 2.5e-4 & 5e-4 \\
& 25\% Shuffled & 5e-5 & 1e-4 & 2.5e-4 & 5e-4 \\
& 0\% Shuffled  & 5e-5 & 1e-4 & 2.5e-4 & 5e-4 \\
\midrule
\multirow{4}{*}{Resnet20/CIFAR100} & $\kappa=1$ & 0.5 & 0.5 & 0.5 & 0.5 \\
& $\kappa=0.1$ & 0.5 & 0.5 & 0.5 & 0.5 \\
& $\kappa=0.05$ & 0.5 & 0.5 & 0.5 & 0.5 \\
& $\kappa=0.01$ & 0.5 & 0.5 & 0.5 & 0.5 \\
\bottomrule
\end{tabular}
\end{center}
\end{table}

\begin{table}[ht]
\small
\caption{DeTAG-specific hyperparameters used for each experiments.}
\label{table:hyperparameter:detag}
\small
\begin{center}
\begin{tabular}{ccccccc}
\toprule
Experiment & Setting & Accelerate Factor $\eta$ & Phase Length $R$\\
\midrule
\multirow{4}{*}{LeNet/CIFAR10} & 100\% Shuffled & 0 & 1 \\
& 50\% Shuffled & 0.2 & 2 \\
& 25\% Shuffled & 0.2 & 2 \\
& 0\% Shuffled & 0.2 & 2 \\
\midrule
\multirow{4}{*}{Resnet20/CIFAR100} & $\kappa=1$ & 0 & 1 \\
& $\kappa=0.1$ & 0.2 & 2 \\
& $\kappa=0.05$ & 0.2 & 2 \\
& $\kappa=0.01$ & 0.4 & 2 \\
\bottomrule
\end{tabular}
\end{center}
\end{table}
\newpage
\section{Technical Proof}
\subsection{Proof to Theorem~\ref{thmlowerbound}}
\begin{proof}
To prove this theorem, it suffices for us to provide two examples, each has a (set of) loss function $f\in\mathcal{F}_{\Delta, L}$, a set of underlying oracles $O\in\mathcal{O}_{\sigma^2}$, a graph $G\in\mathcal{G}_{n,D}$, such that $\inf_{A\in\mathcal{A}_B}T_\epsilon(A, f, O, G)$ is lower bounded by $\Omega\left(\frac{\Delta L\sigma^2}{nB\epsilon^4}\right)$ and $\Omega\left(\frac{\Delta LD}{\epsilon^2}\right)$ iterations on these two examples, respectively. 
Then we will obtain the final bound as $\max\left\{\Omega\left(\frac{\Delta L\sigma^2}{nB\epsilon^4}\right), \Omega\left(\frac{\Delta LD}{\epsilon^2}\right)\right\}$, i.e., $\Omega\left(\frac{\Delta L\sigma^2}{nB\epsilon^4} + \frac{\Delta LD}{\epsilon^2}\right)$ as desired. For simplicity, we denote $\*z^{(i)}$ as the $i$-th coordinate of vector $\*z\in\mathbb{R}^d$.

For each setting, our constructions contain three main steps. 

(1) The first step is to follow the construction of a zero chain function model \cite{carmon2017lower,carmon2019lower}. 
Following \cite{arjevani2019lower} and define
\begin{equation}
    \text{prog}(\*z) = \max\{i\geq 0 | \*z^{(i)}\neq0\}, \forall \*z\in\mathbb{R}^d.
\end{equation}
A zero chain function $f$ has the following property:
\begin{equation}
    \text{prog}(\nabla f(\*x)) \leq \text{prog}(\*x) + 1,
\end{equation}
that means, for a model start from $\*x=\*0$, a single gradient evaluation can only make at most one more coordinate to be non-zero. The name of "chain" comes from the fact that the adjacent coordinates are linked like a chain and only if the previous coordinate becomes non-zero that the current coordinate can become non-zero via a gradient update. Consider a model with $d$ dimension, if we show that $\|\nabla f(\*x)\|\geq \epsilon$ for any $\*x\in\mathbb{R}^d$ with $\*x^{(d)}=0$, we will obtain $d$ as a lower bound on the gradient calls to obtain the $\epsilon$-stationary point. We refer such sequential lower bound as $T_0$.

(2) Step two is to construct a graph $G\in\mathcal{G}_{n,D}$ and a set of oracle $O\in\mathcal{O}_{\sigma^2}$. To do this, our basic idea is to follow \cite{arjevani2019lower} and introduce randomness on the $\text{prog}(\*x)$, and thus the whole chain only make progress with probability $p$. As will be shown later, this requires $\Omega(T_0/p)$ iterations in total.

(3) The third and last step is to rescale the function and distribution so as to make it belong to the function and oracle classes we consider. In other words, this step is to guarantee the result is shown in terms of $\Delta$, $L$, $\sigma$, $n$ and $D$.

We start from a smooth and (potentially) non-convex zero chain function $\hat{f}$ \cite{carmon2019lower} as defined below:
\begin{equation}\label{Equation base zero chain}
    \hat{f}(\*x) = -\Psi(1)\Phi(\*x^{(1)}) + \sum_{i=1}^{T-1}[\Psi(-\*x^{(i)})\Phi(-\*x^{(i+1)}) - \Psi(\*x^{(i)}\Phi(\*x^{(i+1)})],
\end{equation}
where for $\forall z\in\mathbb{R}$
\begin{equation}
    \Psi(z) = \left\{
    \begin{array}{ll}
        0 & \quad z \leq 1/2 \\
        \text{exp}\left(1-\frac{1}{(2z-1)^2}\right) & \quad z > 1/2
    \end{array}
    \right. , \hspace{4em}
    \Phi(z) = \sqrt{e}\int_{-\infty}^{z}e^{\frac{1}{2}t^2}dt.
\end{equation}
This function, as shown in previous works \cite{carmon2019lower,arjevani2019lower}, is a zero-chain function and thus is generally "hard" to optimize: it costs at least $T$ gradient evaluations to find a stationary point. We summarize some properties of Equation~(\ref{Equation base zero chain}) as the following (Proof can be found in Lemma 2 in \cite{arjevani2019lower}):
\begin{enumerate}
    \item \label{zerochainpropertyvalue} $\hat{f}(\*x) - \inf_x\hat{f}(\*x) \leq \Delta_0T$, $\forall \*x\in\mathbb{R}^d$, where $\Delta_0=12$.
    \item \label{zerochainpropertylipschitz}
    $\hat{f}$ is $l_1$-smooth, where $l_1=152$.
    \item \label{zerochainpropertyinfinitynorm}
    $\forall \*x\in\mathbb{R}^T$, $\|\nabla\hat{f}(\*x)\|_\infty\leq G_\infty$, where $G_\infty=23$.
    \item \label{zerochainpropertygradient} $\forall \*x\in\mathbb{R}^T$, if $\text{prog}(\*x)<T$, then $\|\hat{f}(\*x)\|_\infty \geq 1$.
\end{enumerate}
\textbf{(Setting 1)}
Next we discuss the first setting with lower bound $\Omega\left(\frac{\Delta L\sigma^2}{nB\epsilon^4}\right)$. (Setting 1, Step 1) The loss functions are defined as
\begin{equation}
\hat{f}_i(\*x) = \hat{f}(\*x),
\end{equation}
note that $1/n\sum_{i=1}^{n}\hat{f}_i = \hat{f}$. It can be seen from Property~\ref{zerochainpropertylipschitz} that all the $\hat{f}_i$ are $l_1$-smooth. (Setting 1, Step 2) For this setting we consider complete graph. We construct the oracle on worker $i$ as the following:
\begin{equation}
    [\hat{g}_i(\*x)]_j = \nabla_j\hat{f}_i(\*x)\cdot\left(1+\mathbbm{1}\{j>\text{prog}(\*x)\}\left(\frac{z}{p}-1\right)\right),
\end{equation}
where $z\sim\text{Bernoulli}(p)$.
It can be seen that
\begin{equation}
    \mathbb{E}[\hat{g}_i(\*x)] = \nabla \hat{f}_i(\*x),
\end{equation}
and from Property~\ref{zerochainpropertyinfinitynorm} we know
\begin{align*}
    \mathbb{E}\|\hat{g}_i(\*x) - \nabla\hat{f}_i(\*x)\|^2 = |\nabla_{\text{prog}(\*x)+1}\hat{f}(\*x)|^2\mathbb{E}\left(\frac{z}{p}-1\right)^2 \leq \frac{\|\nabla \hat{f}_i(\*x)\|_\infty^2(1-p)}{p} \leq \frac{\|\nabla \hat{f}(\*x)\|_\infty^2(1-p)}{p} \leq \frac{G_\infty^2(1-p)}{p}.
\end{align*}
(Setting 1, Step 3) Finally we rescale each function as $f_i = L\lambda^2/l_1\hat{f}_i(\*x/\lambda)$ where $\lambda$ is a parameter subject to change. For $L$: note that all $f_i$ are $\frac{L}{l_1}\cdot l_1=L$-smooth. For the $\Delta$, 
\begin{equation}\label{equationsetting1delta}
    f - f^* = \frac{L\lambda^2}{l_1}(\hat{f}-\hat{f}^*) = \frac{L\lambda^2\Delta_0T}{l_1} \leq \Delta.
\end{equation}
For the oracle, to be consistent with $f_i$, we rescale it as $g_i(\*x) = L\lambda/l_1\hat{g}_i(\*x/\lambda)$, and we have
\begin{equation}\label{equationsetting1sigma}
    \mathbb{E}\|g_i(\*x)-\nabla f_i(\*x)\|^2 \leq \frac{L^2\lambda^2}{l_1^2} \mathbb{E}\left\|g_i\left(\frac{\*x}{\lambda}\right)-\nabla f_i\left(\frac{\*x}{\lambda}\right)\right\|^2 \leq \frac{L^2\lambda^2G_\infty^2(1-p)}{l_1^2p} \leq \sigma^2.
\end{equation}
We assign $\lambda=2l_1\epsilon/L$, then Equation~(\ref{equationsetting1delta}) and (\ref{equationsetting1sigma}) are fulfilled with 
\begin{align*}
    T = & \left\lfloor \frac{\Delta}{\Delta_0l_1(2\epsilon)^2}\right\rfloor, \\
    p = & \min\{(2G_\infty\epsilon)^2/\sigma^2, 1\}.
\end{align*}
Take $\delta=1/2$ in Lemma~\ref{lemma prob chain}, we have for probability at least 1/2, $\|\nabla f(\*{\hat{x}}^{(t)})\|\geq \epsilon$ for all $t\leq\frac{T+\log(\delta)}{\min\{nBp, 1\}(e-1)}$. Use Property~\ref{zerochainpropertygradient}, for any $\*x\in\mathbb{R}^T$ such that $\text{prog}(\*x)<T$ it holds that $\|\nabla f(\*x)\|\geq 2\epsilon$, therefore,
\begin{equation}
    \mathbb{E}\|\nabla f(\*{\hat{x}}_T)\| >\epsilon.
\end{equation}
Then with small $\epsilon$ it follows that
\begin{equation}
    T_\epsilon(A, f, O, G) \geq \frac{T-1}{nBp(e-1)} \geq \Omega\left(\frac{\Delta L\sigma^2}{nB\epsilon^4}\right),
\end{equation}
and that completes the proof for setting 1.

\begin{figure}[t]
    \centering
    \includegraphics[width=0.5\linewidth]{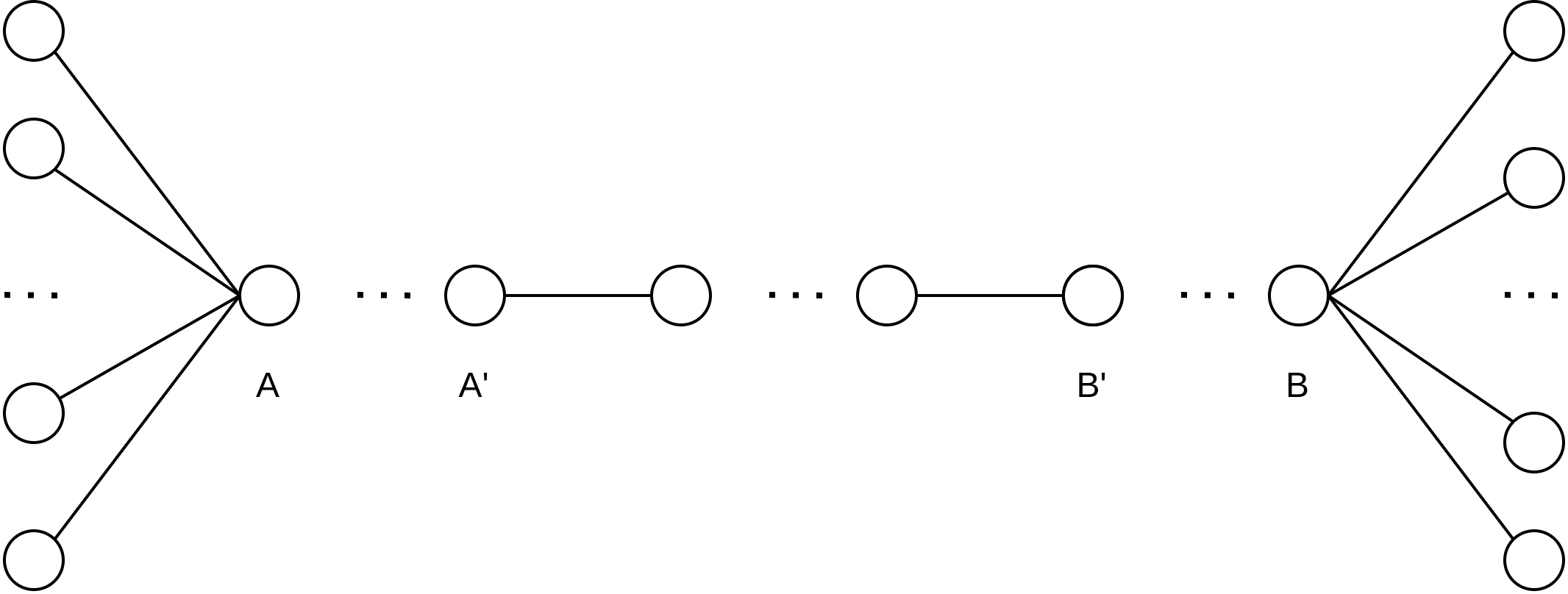}
    \caption{Illustration graph for setting 2 to in the proof of Theorem~\ref{thmlowerbound}.}
    \label{fig:graph}
\end{figure}

\textbf{(Setting 2)}
We proceed to the prove second bound $\Omega\left(\frac{\Delta LD}{\epsilon^2}\right)$.

(Setting 2 Step 1 \& Step 2) 
We assign all the workers with index from $1$ to $n$, we first define two indices set
\begin{equation}
\begin{aligned}
I_0 = & \left\{1, \cdots, |I_0| \right\}, \\
I_1 = & \left\{n, n-1, \cdots, n-|I_1|+1 \right\}.
\end{aligned}
\end{equation}
where $|\cdot|$ denotes a cardinality of a set. Consider the construction of $G$ in Figure~\ref{fig:graph}:

If $D\geq n-2\lceil n/3\rceil+2$, then it implies the number of nodes between A and B is larger than $\lceil n/3\rceil$. In this case, denote A' and B' as a sub linear graph where its number of nodes is exactly $\lceil n/3\rceil$. Let all the nodes on the left of A' be in $I_0$ and all the nodes on the right of B' be $I_1$
We define all the local functions on such graph as following:
\begin{equation}
\hat{f}_i(\*x) = \left\{
\begin{array}{ll}
    -\frac{2n}{n-\lceil n/3\rceil}\Psi(1)\Phi(\*x^{(1)}) + \sum_{i=2k, k\in\{1,2, \cdots\},i<T}\frac{2n}{n-\lceil n/3\rceil}[\Psi(-\*x^{(i)})\Phi(-\*x^{(i+1)}) - \Psi(\*x^{(i)})\Phi(\*x^{(i+1)})] & \quad i\in I_0, \\
    \\
    \sum_{i=2k-1, k\in\{1,2, \cdots\},i<T}\frac{2n}{n-\lceil n/3\rceil}[\Psi(-\*x^{(i)})\Phi(-\*x^{(i+1)}) - \Psi(\*x^{(i)})\Phi(\*x^{(i+1)})] & \quad i\in I_1, \\
    \\
    0 & \quad i\not\in I_0,I_1. \\
\end{array}
\right.
\end{equation}

If $D< n-2\lceil n/3\rceil+2$, the distance between node A and node B is $D-2$ and the sub linear graph whose end points are A and B contains $D-1$ nodes. We let the number of nodes on the left of $A$ be $\left\lceil\frac{n-D+1}{2}\right\rceil$, we denote the set of indices of all such nodes as $I_0$; and then we let the number of nodes on the right of $B$ be $\left\lfloor\frac{n-D+1}{2}\right\rfloor$, we denote the set of indices of all such nodes as $I_I$.
Since $D< n-2\lceil n/3\rceil+2$, this implies $|I_0|, |I_1|>n/3$.
We define all the local functions on such graph as following:
\begin{equation}
\hat{f}_i(\*x) = \left\{
\begin{array}{ll}
    -\frac{n}{|I_0|}\Psi(1)\Phi(\*x^{(1)}) + \sum_{i=2k, k\in\{1,2, \cdots\},i<T}\frac{n}{|I_0|}[\Psi(-\*x^{(i)})\Phi(-\*x^{(i+1)}) - \Psi(\*x^{(i)})\Phi(\*x^{(i+1)})] & \quad i\in I_0, \\
    \\
    \sum_{i=2k-1, k\in\{1,2, \cdots\},i<T}\frac{n}{|I_1|}[\Psi(-\*x^{(i)})\Phi(-\*x^{(i+1)}) - \Psi(\*x^{(i)})\Phi(\*x^{(i+1)})] & \quad i\in I_1, \\
    \\
    0 & \quad i\not\in I_0,I_1. \\
\end{array}
\right.
\end{equation}
In both cases discussed based on $D$,
we can see that $\hat{f}(\*x) = \frac{1}{n}\sum_{i=1}^{n}\hat{f}_i(\*x)$, and we are splitting hard zero-chain function into two main different part: the even components of the chain and the odd components of the chain. 
It is easy to see that for the zero chain function to make progress, it takes at least $\lceil n/3\rceil$, i.e., $\Omega(D)$ number of iterations in the first case (since here $D=\tilde{\gamma}n$ for some $\tilde{\gamma}>1/3$) and $D$ number of iterations in the seconds case. Then the total number of iterations is lower bounded by $\Omega(TD)$.

For the oracle, we let oracle on worker $i$ as
\begin{equation}
    [\hat{g}_i(\*x)]_j = \nabla_j\hat{f}_i(\*x).
\end{equation}
(Setting 2, Step 3)
The last step is to rescale the parameters. Compared to setting 1, we know here all the $\hat{f}_i$ are $3l_1$-smooth, as before we let
\begin{equation}
    f_i(\*x) = \frac{L\lambda^2}{3l_1}\hat{f}_i\left(\frac{\*x}{\lambda}\right), \hspace{2em}\lambda=\frac{6l_1\epsilon}{L}.
\end{equation}
For the $\Delta$ bound we have
\begin{equation}
    L\lambda^2\Delta_0T/3l_1\leq \Delta
\end{equation}
to fulfill this it suffices to set
\begin{equation}
    T=\left\lfloor\frac{\Delta L}{\Delta_0l_1(12\epsilon)^2}\right\rfloor.
\end{equation}
It also can be seen that $f$ is $L$-smooth. So in this setting,
\begin{equation}
    T_\epsilon(A, f, O, G) \geq \Omega(TD) = \Omega\left(\frac{\Delta LD}{\epsilon^2}\right).
\end{equation}
Combining Setting 1 and 2 we complete the proof.
\end{proof}

\begin{lemma}\label{lemma prob chain}
In setting 1 in the proof of Theorem 1, with probability at least $1-\delta$, $\|\nabla f(\*x_t)\|\geq \epsilon$ for all $t\leq\frac{T+\log(\delta)}{\min\{nBp, 1\}(e-1)}$.
\end{lemma}
\begin{proof}
    Define a filtration at iteration $t$ as the sigma field of all the previous events happened before iteration $t$.
    Let $i^{(t)}_j=\text{prog}(\*x_{t,j}), \forall j \in[n]$ and $i^{(t)} = \max_j{i^{(t)}_j}$. And we denote $\mathcal{E}^{(t,m,j)}$ as the event of \texttt{the $i^{(t)}_m + 1$-th coordinate of output of $j$-th query on worker $m$ at iteration $t$ is non-zero}. Based on the independent sampling, these events are independent. Thus we know:
    \begin{equation}
        \mathbb{P}[i^{(t+1)} - i^{(t)} = 1 | \mathcal{U}^{(t)}] = \mathbb{P}\left[\bigcup_{\substack{i\in[n]\\ j\leq B}}\mathcal{E}^{(t,i,j)} | \mathcal{U}^{(t)}\right] \leq \sum_{i\in[n],j\leq B}\mathbb{P}\left[\mathcal{E}^{(t,i,j)} | \mathcal{U}^{(t)}\right] \leq \min\{nBp, 1\}.
    \end{equation}
    Let $q^{(t)} = i^{(t+1)} - i^{(t)}$, with Chernoff bound, we obtain
    \begin{equation}
        \mathbb{P}[i^{(t)} \geq T] = \mathbb{P}[e^{\sum_{j=0}^{t-1}q^{(j)}} \geq e^T] \leq e^{-T}\mathbb{E}[e^{\sum_{j=0}^{t-1}q^{(j)}}].
    \end{equation}
    For the expectation term we know that
    \begin{equation}
        \mathbb{E}[e^{\sum_{j=0}^{t-1}q^{(j)}}] = \mathbb{E}\left[\prod_{j=0}^{t-1}\mathbb{E}\left[e^{q^{(j)}}|\mathcal{U}^{(j)}\right]\right] \leq (1-\min\{nBp, 1\}+\min\{nBp, 1\}e)^t \leq e^{\min\{nBp, 1\}t(e-1)}.
    \end{equation}
    Thus we know
    \begin{equation}
        \mathbb{P}[i^{(t)}\geq T] \leq e^{(e-1)\min\{nBp, 1\}t-T} \leq \delta,
    \end{equation}
    for every $t\leq \frac{T+\log(\delta)}{\min\{nBp, 1\}(e-1)}$.
\end{proof}

\subsection{Proof to Corollary~\ref{corollary lower bound}}
\begin{proof}
Different from Theorem~\ref{thmrateDeTAG}, in this corollary we do not choose $D$ and $n$ separately, so that our construction can just use the linear graph as follows:

(Linear graph, Step 1) We first let $|I_0|=|I_1|=\lceil n/3\rceil$ in the proof of Theorem~\ref{thmlowerbound}, meaning $I_0$ denotes the first $\lceil n/3\rceil$ workers and $I_1$ denotes the last $\lceil n/3\rceil$ workers. We define all the local functions $\hat{f}_i(\*x)$ as following:
\begin{equation}
\left\{
\begin{array}{ll}
    -\frac{n}{\lceil n/3\rceil}\Psi(1)\Phi(\*x^{(1)}) + \sum_{i=2k, k\in\{1,2, \cdots\},i<T}\frac{n}{\lceil n/3\rceil}[\Psi(-\*x^{(i)})\Phi(-\*x^{(i+1)}) - \Psi(\*x^{(i)})\Phi(\*x^{(i+1)})] & \quad i\in I_0, \\
    \\
    \sum_{i=2k-1, k\in\{1,2, \cdots\},i<T}\frac{n}{\lceil n/3\rceil}[\Psi(-\*x^{(i)})\Phi(-\*x^{(i+1)}) - \Psi(\*x^{(i)})\Phi(\*x^{(i+1)})] & \quad i\in I_1, \\
    \\
    0 & \quad i\not\in I_0,I_1. \\
\end{array}
\right.
\end{equation}
we can see that $\hat{f}(\*x) = \frac{1}{n}\sum_{i=1}^{n}\hat{f}_i(\*x)$. 
(Linear graph, Step 2)
We consider linear graph in this setting and from one end to the other, the worker's index is $1$ to $n$, without the loss of generality.
It is easy to see that for the zero chain function to make progress, it takes at least $n-2\lceil n/3\rceil+1$ number of iterations. Note that in linear graph $n-1=D$, the total number of iterations is at least
\begin{equation}
    \Omega\left(TD\right).
\end{equation}
For the oracle, we let oracle on worker $i$ as
\begin{equation}
    [\hat{g}_i(\*x)]_j = \nabla_j\hat{f}_i(\*x)
\end{equation}
(Linear graph, Step 3)
The last step is to rescale the parameters. Compared to setting 1, we know here all the $\hat{f}_i$ are $3l_1$-smooth, as before we let
\begin{equation}
    f_i(\*x) = \frac{L\lambda^2}{3l_1}\hat{f}_i\left(\frac{\*x}{\lambda}\right), \hspace{2em}\lambda=\frac{6l_1\epsilon}{L}.
\end{equation}
For the $\Delta$ bound we have
\begin{equation}
    L\lambda^2\Delta_0T/3l_1\leq \Delta,
\end{equation}
to fulfill this it suffices to set
\begin{equation}
    T=\left\lfloor\frac{\Delta L}{\Delta_0l_1(12\epsilon)^2}\right\rfloor.
\end{equation}
It also can be seen that $f$ is $L$-smooth. So in this setting,
\begin{equation}
    T_\epsilon(A, f, O, G) \geq \Omega(TD) \geq \Omega\left(\frac{\Delta LD}{\epsilon^2}\right).
\end{equation}
Given the bound, we use two additional results on linear graph as \cite{berthier2020accelerated}: the random walk matrix $\*W_{rw}$ on linear graph with $\lambda$ fulfilling
\begin{align}
    \frac{1}{\sqrt{1-\lambda}}=O(D).
\end{align}
Then we can rewrite the lower bound in the form of $\lambda$ as shown in Corollary~\ref{corollary lower bound}.

Finally, using the conclusion of $\lambda=\cos(\pi/n)$ for $n\in\{2, 3, \cdots, \}$ on linear graph we complete the proof.

\end{proof}
\subsection{Proof to Theorem~\ref{thmrateDeFacto}}
\begin{proof}
As (partially) discussed in the paper, DeFacto is statistically equivalent to centralized SGD. 
Specifically, it conduct $K=T/2R$ gradient steps where each step contains a mini-batch of $R$ at the point of $\*x_{k,i}, \forall i\in[n]$. Take the well-known convergence rate for centralized SGD:
\begin{equation}
    \frac{1}{T}\sum_{t=0}^{T-1}\|\nabla f(\*{\hat{x}})\|^2 \leq O\left(\frac{\Delta L\sigma}{\sqrt{nBT}}+\frac{\Delta L}{T}\right).
\end{equation}
The convergence rate of DeFacto can be expressed as:
\begin{equation}
    \frac{1}{T}\sum_{t=0}^{T-1}\|\nabla f(\*{\hat{x}})\|^2 \leq O\left(\frac{\Delta L\sigma/\sqrt{R}}{\sqrt{nBK}}+\frac{\Delta L}{K}\right) = O\left(\frac{\Delta L\sigma}{\sqrt{nBT}}+\frac{\Delta LR}{T}\right) = O\left(\frac{\Delta L\sigma}{\sqrt{nBT}}+\frac{\Delta LD}{T}\right),
\end{equation}
then we obtain for DeFacto, when $T=O(\Delta L\sigma^2(nB\epsilon^4)^{-1} + \Delta LD\epsilon^{-2})$,
\begin{equation}
    \min_{t=0,1,\cdots, T-1}\mathbb{E}\left\|\nabla f\left(\*{\hat{x}}\right)\right\| \leq \sqrt{\min_{t=0,1,\cdots, T-1}\mathbb{E}\left\|\nabla f\left(\*{\hat{x}}\right)\right\|^2} \leq \sqrt{O\left(\frac{\Delta L\sigma}{\sqrt{nBT}} + \frac{\Delta LD}{T}\right)}\leq \epsilon,
\end{equation}
that completes the proof.
\end{proof}
\subsection{Proof to Theorem~\ref{thmrateDeTAG}}
\begin{proof}
In this proof, we adopt an updated version of notation: we denote at the beginning of phase $k$, the three quantities of interests are $\*X_k$, $\*Y_k$ and $\tilde{\*G}_k$, and the update rule becomes:
\begin{align}
    & \*Y_{k+1} = \mathcal{M}(\*Y_k+\tilde{\*G}_{k} - \tilde{\*G}_{k-1}), \\
    & \*X_{k+1} = \mathcal{M}(\*X_k-\alpha \*Y_k),
\end{align}
with
\begin{align}
    \tilde{\*G}_{k+1} = & \left[\nabla\tilde{f}_1(\*x_{k,1}), \cdots, \nabla\tilde{f}_n(\*x_{k,n})\right]\in\mathbb{R}^{d\times n}, \\
    \*G_{k+1} = & \left[\nabla f_1(\*x_{k,1}), \cdots, \nabla f_n(\*x_{k,n})\right]\in\mathbb{R}^{d\times n}, \\
    {\*X}_{k} = & \left[\*x_{k,1}, \cdots, \*x_{k,n}\right]\in\mathbb{R}^{d\times n}, \\
    {\*Y}_{k} = & \left[\*y_{k,1}, \cdots, \*y_{k,n}\right]\in\mathbb{R}^{d\times n}, 
\end{align}
where $\nabla\tilde{f}_i$ denotes the stochastic gradient oracle on worker $i$, and  $\nabla f_i$ denotes the full gradient oracle on worker $i$.
We use $\overline{\*X}$ denote $\*X\frac{\*1}{n}$ for any matrix $\*X$ with appropriate shape. 
We use $\lambda_i(\*W)$ to denote the $i$-th general  largest eigenvalue of matrix $\*W$. 
Under such notation, $\lambda$ in the main paper is equavilent to $\lambda_2(\*W)$.
We use $\mathcal{M}(\cdot)$ to denote the $R$-step accelerated gossip which has the following property \cite{liu2011accelerated}:
\begin{align}\label{proof equation property AG}
    \|\mathcal{M}(\*X)-\overline{\*X}\| \leq  \rho \|\*X-\overline{\*X}\|; \hspace{1em} \mathcal{M}(\*X)\frac{\*1}{n}=\*X\frac{\*1}{n},
\end{align}
where $\rho =\left(1-\sqrt{1-\lambda_2(\*W)}\right)^R$. The proof to the statement of Equation~(\ref{proof equation property AG}) can be found in \cite{ye2020multi}.

For the stochastic oracle, based on the oracle class assumption, we have
\begin{align}
    \mathbb{E}\|\nabla\tilde{f}_i(\*x)-\nabla f_i(\*x)\|^2\leq\sigma^2,
\end{align}
and we denote $\tilde{\sigma}^2=\frac{\sigma^2}{BR}$ as the variance of mini-batch of $R$.

First, from the update rule of DeTAG,
\begin{equation}\label{proof equation Yk}
    \overline{\*Y}_{k} = \mathcal{M}(\*Y_{k-1} + \tilde{\*G}_{k-1} - \tilde{\*G}_{k-2})\frac{\*1}{n} = \overline{\*Y}_{k-1} + \overline{\tilde{\*G}}_{k-1} - \overline{\tilde{\*G}}_{k-2} = \overline{\*Y}_{-1} + \sum_{j=-1}^{k-1}(\overline{\tilde{\*G}}_{j} - \overline{\tilde{\*G}}_{j-1}) = \overline{\tilde{\*G}}_{k-1}
\end{equation}
and
\begin{equation}\label{proof equation Xk}                       \overline{\*X}_{k+1} = \mathcal{M}(\*X_k-\alpha\*Y_k)\frac{\*1}{n} = \overline{\*X}_{k} - \alpha\overline{\*Y}_{k}.
\end{equation}
By Taylor Theorem, we obtain
\begin{align}
    \mathbb{E}f\left(\overline{\*X}_{k+1}\right)
    = & \mathbb{E}f\left(\overline{\*X}_k - \alpha\overline{\*Y}_{k}\right)\\
    \leq & \mathbb{E}f\left(\overline{\*X}_k\right)-\alpha\mathbb{E}\left\langle \nabla f\left(\overline{\*X}_k\right), \overline{\*Y}_{k}\right\rangle + \frac{\alpha^2L}{2}\mathbb{E}\left\|\overline{\*Y}_{k}\right\|^2\\
    \overset{(\ref{proof equation Yk})}{=} & \mathbb{E}f\left(\overline{\*X}_k\right)-\alpha\mathbb{E}\left\langle \nabla f\left(\overline{\*X}_k\right), \overline{\*G}_{k-1}\right\rangle + \frac{\alpha^2L}{2}\mathbb{E}\left\|\overline{\tilde{\*G}}_{k-1}\right\|^2.
\end{align}
For the last term, we have
\begin{align}\label{proof equation sampling noise}
    \mathbb{E}\left\|\overline{\tilde{\*G}}_{k-1}\right\|^2
    = & \mathbb{E}\left\|\overline{\*G}_{k-1}\right\|^2 + \mathbb{E}\left\|\overline{\*G}_{k-1}-\overline{\tilde{\*G}}_{k-1}\right\|^2 + 2\mathbb{E}\left\langle \overline{\*G}_{k-1},\overline{\*G}_{k-1}-\overline{\tilde{\*G}}_{k-1}\right\rangle\\
    = & \mathbb{E}\left\|\overline{\*G}_{k-1}\right\|^2 + \mathbb{E}\left\|\overline{\*G}_{k-1}-\overline{\tilde{\*G}}_{k-1}\right\|^2\\
    = & \mathbb{E}\left\|\overline{\*G}_{k-1}\right\|^2 + \frac{1}{n^2}\sum_{i=1}^{n}\mathbb{E}\left\|\*G_{k-1}\*e_i-\*{\tilde{G}}_{k-1}\*e_i\right\|^2\\
    \leq & \mathbb{E}\left\|\overline{\*G}_{k-1}\right\|^2 + \frac{\tilde{\sigma}^2}{n},
\end{align}
where in the second step, we use the fact that the sampling noise is independent of the gradient itself.
Putting it back we obtain
\begin{align}
    \mathbb{E}f\left(\overline{\*X}_{k+1}\right)
    \leq & \mathbb{E}f\left(\overline{\*X}_k\right)-\alpha\mathbb{E}\left\langle \nabla f\left(\overline{\*X}_k\right), \overline{\*G}_{k-1}\right\rangle + \frac{\alpha^2L}{2}\mathbb{E}\left\|\overline{\*G}_{k-1}\right\|^2 + \frac{\alpha^2\tilde{\sigma}^2L}{2n}\\
    = & \mathbb{E}f\left(\overline{\*X}_k\right)-\frac{\alpha}{2}\mathbb{E}\left\|\nabla f\left(\overline{\*X}_k\right)\right\|^2 - \frac{\alpha-\alpha^2L}{2}\left\|\overline{\*G}_{k-1}\right\|^2 + \frac{\alpha^2\tilde{\sigma}^2L}{2n} + \frac{\alpha}{2}\mathbb{E}\left\|\overline{\*G}_{k-1} - \nabla f\left(\overline{\*X}_k\right)\right\|^2,
\end{align}
where the last step we use $2\langle a, b\rangle = \|a\|^2 + \|b\|^2 - \|a-b\|^2$.
Expand the last term, we obtain
\begin{align}
    & \mathbb{E}\left\|\overline{\*G}_{k-1} - \nabla f\left(\overline{\*X}_k\right)\right\|^2\\
    \leq & 2\mathbb{E}\left\|\overline{\*G}_{k-1} - \overline{\*G}_{k+1}\right\|^2 + 2\mathbb{E}\left\|\overline{\*G}_{k+1}-\nabla f\left(\overline{\*X}_k\right)\right\|^2\\
        = & 2\mathbb{E}\left\|\frac{1}{n}\sum_{i=1}^{n}\nabla f_i(\*x_{k,i}) - \frac{1}{n}\sum_{i=1}^{n}\nabla f_i(\*x_{k-2,i})\right\|^2 + 2\mathbb{E}\left\|\frac{1}{n}\sum_{i=1}^{n}\nabla f_i(\*x_{k,i})-\frac{1}{n}\sum_{i=1}^{n}\nabla f_i(\overline{\*X}_k)\right\|^2\\
    \leq  & \frac{2}{n}\sum_{i=1}^{n}\mathbb{E}\left\|\nabla f_i(\*x_{k,i}) - \nabla f_i(\*x_{k-2,i})\right\|^2 + \frac{2}{n}\sum_{i=1}^{n}\mathbb{E}\left\|\nabla f_i(\*x_{k,i})-\nabla f_i(\overline{\*X}_k)\right\|^2\\
        \leq & \frac{2L^2}{n}\mathbb{E}\left\|\*X_k - \*X_{k-2}\right\|_F^2 + \frac{2L^2}{n}\mathbb{E}\left\|\*X_k-\overline{\*X}_k\*{1}_n^\top\right\|_F^2.
\end{align}
Denote $f(\*0)-f^*\leq \Delta$, we obtain
\begin{align}
    & \sum_{k=0}^{K-1}\alpha(1-\alpha L)\left\|\overline{\*G}_k\right\|^2 + \sum_{k=0}^{K-1}\alpha\mathbb{E}\left\|\nabla f\left(\overline{\*X}_k\right)\right\|^2\\
    \leq & 2\Delta + \frac{\alpha^2\tilde{\sigma}^2LK}{n} + \frac{2\alpha L^2}{n}\sum_{k=0}^{K-1}\mathbb{E}\left\|\*X_k-\overline{\*X}_k\*{1}_n^\top\right\|_F^2 + \frac{2\alpha L^2}{n}\sum_{k=0}^{K-1}\mathbb{E}\left\|\*X_k - \*X_{k-2}\right\|_F^2\\
    \leq & 2\Delta + \frac{\alpha^2\tilde{\sigma}^2LK}{n} + \frac{16\alpha L^2}{n}\sum_{k=0}^{K-1}\mathbb{E}\left\|\*X_k-\overline{\*X}_k\*{1}_n^\top\right\|_F^2 + \frac{6\alpha L^2}{n}\sum_{k=0}^{K-1}\mathbb{E}\left\|\overline{\*X}_k\*{1}_n^\top - \overline{\*X}_{k-2}\*{1}_n^\top\right\|_F^2,
\end{align}
where in the last step we use
\begin{align}
    & \frac{2\alpha L^2}{n}\sum_{k=0}^{K-1}\mathbb{E}\left\|\*X_k - \*X_{k-2}\right\|_F^2\\
    \leq & \frac{6\alpha L^2}{n}\sum_{k=0}^{K-1}\mathbb{E}\left\|\*X_k - \overline{\*X}_k\*{1}_n^\top\right\|_F^2
    + \frac{6\alpha L^2}{n}\sum_{k=0}^{K-1}\mathbb{E}\left\|\*X_{k-2} - \overline{\*X}_{k-2}\*{1}_n^\top\right\|_F^2
    + \frac{6\alpha L^2}{n}\sum_{k=0}^{K-1}\mathbb{E}\left\|\overline{\*X}_k\*{1}_n^\top - \overline{\*X}_{k-2}\*{1}_n^\top\right\|_F^2.
\end{align}
In addition, for the last term we have
\begin{align}
    \frac{6\alpha L^2}{n}\sum_{k=0}^{K-1}\mathbb{E}\left\|\overline{\*X}_k\*{1}_n^\top - \overline{\*X}_{k-2}\*{1}_n^\top\right\|_F^2 = &  \frac{6\alpha L^2n}{n}\sum_{k=0}^{K-1}\mathbb{E}\left\|\overline{\*X}_k - \overline{\*X}_{k-2}\right\|^2\\
    \overset{(\ref{proof equation Xk})}{=} & \frac{24\alpha^3 L^2n}{n}\sum_{k=0}^{K-1}\mathbb{E}\left\|\overline{\tilde{\*G}}_{k}\right\|^2\\
    \overset{(\ref{proof equation sampling noise})}{\leq} & 24\alpha^3 L^2\sum_{k=0}^{K-1}\mathbb{E}\left\|\overline{{\*G}}_{k}\right\|^2 + \frac{24\alpha^3\tilde{\sigma}^2L^2K}{n}.
\end{align}
Push it back we have
\begin{align}\label{proof equation main formula}
    & \sum_{k=0}^{K-1}\alpha(1-\alpha L-24\alpha^2L^2)\left\|\overline{\*G}_k\right\|^2 + \sum_{k=0}^{K-1}\alpha\mathbb{E}\left\|\nabla f\left(\overline{\*X}_k\right)\right\|^2\\
    \leq & 2\Delta + \frac{\alpha^2\tilde{\sigma}^2LK}{n} + \frac{16\alpha L^2}{n}\sum_{k=0}^{K-1}\mathbb{E}\left\|\*X_k-\overline{\*X}_k\*{1}_n^\top\right\|_F^2 + \frac{24\alpha^3\tilde{\sigma}^2L^2K}{n}.
\end{align}
The rest of the proof is to bound $\frac{16\alpha L^2}{n}\sum_{k=0}^{K-1}\mathbb{E}\left\|\*X_k-\overline{\*X}_k\*{1}_n^\top\right\|_F^2$.

We start from
\begin{align}
    & \left\|\*X_{k+1}-\overline{\*X}_{k+1}\*{1}_n^\top\right\|_F^2\\
    \overset{(\ref{proof equation Xk})}{=} & \left\|\mathcal{M}(\*X_k-\alpha \*Y_k)-(\overline{\*X}_k-\alpha\overline{\*Y}_k)\*{1}_n^\top\right\|_F^2\\
    =  & \left\|\mathcal{M}(\*X_k)-\overline{\*X}_k\*{1}_n^\top\right\|_F^2 - 2\alpha\left\langle \mathcal{M}(\*X_k)-\overline{\*X}_k\*{1}_n^\top, \mathcal{M}(\*Y_k)-\overline{\*Y}_k\*{1}_n^\top\right\rangle + \alpha^2\left\|\mathcal{M}(\*Y_k)-\overline{\*Y}_k\*{1}_n^\top\right\|_F^2\\
    \overset{(\ref{proof equation property AG})}{\leq} & \rho ^2\left\|\*X_k-\overline{\*X}_k\*{1}_n^\top\right\|_F^2 + \frac{\rho ^2(1-\rho ^2)}{1+\rho ^2}\left\|\*X_k-\overline{\*X}_k\*{1}_n^\top\right\|_F^2 + \frac{\rho ^2(1+\rho ^2)\alpha^2}{1-\rho ^2}\left\|\*Y_k-\overline{\*Y}_k\*{1}_n^\top\right\|_F^2\\
        & + \alpha^2\rho ^2\left\|\*Y_k-\overline{\*Y}_k\*{1}_n^\top\right\|_F^2\\
    \label{proof equation X-Xbar}
    = & \frac{2\rho ^2}{(1+\rho ^2)}\left\|\*X_k-\overline{\*X}_k\*{1}_n^\top\right\|_F^2 + \frac{2\rho ^2\alpha^2}{1-\rho ^2}\left\|\*Y_k-\overline{\*Y}_k\*{1}_n^\top\right\|_F^2,
\end{align}
where in the third step we use
\begin{align}\label{proof equation square trick}
    -2\langle \*a, \*b\rangle \leq \frac{1-\rho^2}{1+\rho^2}\|\*a\|^2 + \frac{1+\rho^2}{1-\rho^2}\|\*b\|^2.
\end{align}
Similarly, for $\*Y_{k+1}$, we obtain
\begin{align}
    & \mathbb{E}\left\|\*Y_{k+1}-\overline{\*Y}_{k+1}\*{1}_n^\top\right\|_F^2\\
    \overset{(\ref{proof equation Yk})}{=} & \mathbb{E}\left\|\mathcal{M}(\*Y_{k} + \tilde{\*G}_{k} - \tilde{\*G}_{k-1})-(\overline{\*Y}_{k} + \overline{\tilde{\*G}}_{k} - \overline{\tilde{\*G}}_{k-1})\*{1}_n^\top\right\|_F^2\\
        = & \mathbb{E}\left\|\mathcal{M}(\*Y_{k})-\overline{\*Y}_{k}\*{1}_n^\top\right\|_F^2 + \mathbb{E}\left\|\mathcal{M}(\tilde{\*G}_{k}-\tilde{\*G}_{k-1})-(\overline{\tilde{\*G}}_{k}-\overline{\tilde{\*G}}_{k-1})\*{1}_n^\top\right\|_F^2\\
        & + 2\mathbb{E}\left\langle \mathcal{M}(\*Y_{k})-\overline{\*Y}_{k}\*{1}_n^\top, \mathcal{M}(\tilde{\*G}_{k}-\tilde{\*G}_{k-1})-(\overline{\tilde{\*G}}_{k}-\overline{\tilde{\*G}}_{k-1})\*{1}_n^\top\right\rangle\\
    \overset{(\ref{proof equation square trick})(\ref{proof equation property AG})}{\leq} & \rho ^2\mathbb{E}\left\|\*Y_{k}-\overline{\*Y}_{k}\*{1}_n^\top\right\|_F^2 + \rho ^2\mathbb{E}\left\|\*G_{k}-\*G_{k-1}-(\overline{\*G}_{k}-\overline{\*G}_{k-1})\*{1}_n^\top\right\|_F^2\\
    & + \frac{(1-\rho ^2)\rho ^2}{1+\rho ^2}\mathbb{E}\left\|\*Y_{k}-\overline{\*Y}_{k}\*{1}_n^\top\right\|_F^2 + \frac{(1+\rho ^2)\rho ^2}{1-\rho ^2}\mathbb{E}\left\|\*G_{k}-\*G_{k-1}-(\overline{\*G}_{k}-\overline{\*G}_{k-1})\*{1}_n^\top\right\|_F^2\\
    & + 2\rho^2\mathbb{E}\|\*G_{k}-\tilde{\*G}_{k}\|_F^2 + 2\rho^2\mathbb{E}\|\*G_{k-1}-\tilde{\*G}_{k-1}\|_F^2 + 2\rho^2\mathbb{E}\|\overline{\*G_{k}}\*{1}_n^\top-\overline{\tilde{\*G}}_{k}\*{1}_n^\top\|_F^2 + 2\rho^2\mathbb{E}\|\overline{\*G_{k-1}}\*{1}_n^\top-\overline{\tilde{\*G}}_{k-1}\*{1}_n^\top\|_F^2\\
    \label{proof equation Y-Ybar}
        \leq & \frac{2\rho ^2}{1+\rho ^2}\mathbb{E}\left\|\*Y_{k}-\overline{\*Y}_{k}\*{1}_n^\top\right\|_F^2 + \frac{4\rho ^2}{1-\rho ^2}\mathbb{E}\left\|\*G_{k+2}-\*G_{k+1}\right\|_F^2\\
    & + \frac{4\rho ^2}{1-\rho ^2}\mathbb{E}\left\|\*G_{k+2}-\*G_{k+1}-\*G_{k}+\*G_{k-1}\right\|_F^2 + 8n\rho ^2\tilde{\sigma}^2,
\end{align}
where in the last step we use $\|I-\frac{\*1\*1^\top}{n}\|\leq 1$ and $\|\*A\*B\|_F\leq \|\*A\|_F\|\*B\|$.

For the second term, we have
\begin{align}
    & \mathbb{E}\left\|\*G_{k+2}-\*G_{k+1}\right\|_F^2\\
    = & \sum_{i=1}^{n}\mathbb{E}\left\|\nabla f(\*x_{k+1,i})-\nabla f(\*x_{k,i})\right\|^2\\
    \leq & L^2\sum_{i=1}^{n}\mathbb{E}\left\|\*x_{k+1,i}-\*x_{k,i}\right\|^2\\
        = & L^2\mathbb{E}\left\|\*X_{k+1}-\*X_{k}\right\|_F^2\\
    \overset{(\ref{proof equation Xk})}{=} & L^2\mathbb{E}\left\|\mathcal{M}(\*X_{k})-\*X_{k}-\alpha\mathcal{M}(\*Y_{k})\right\|_F^2\\
        = & L^2\mathbb{E}\left\|\mathcal{M}(\*X_{k}-\overline{\*X}_{k}\*{1}_n^\top)-(\*X_{k}-\overline{\*X}_{k}\*{1}_n^\top)-\alpha\mathcal{M}(\*Y_{k})\right\|_F^2\\
    \leq & 4L^2\mathbb{E}\left\|\mathcal{M}(\*X_{k})-\overline{\*X}_{k}\*{1}_n^\top\right\|_F^2 + 4L^2\mathbb{E}\left\|\*X_{k}-\overline{\*X}_{k}\*{1}_n^\top\right\|_F^2 + 4\alpha^2L^2\mathbb{E}\left\|\mathcal{M}(\*Y_{k})-\overline{\*Y}_{k}\*{1}_n^\top\right\|_F^2\\
        & + 4\alpha^2nL^2\mathbb{E}\left\|\overline{\*Y}_{k}\right\|^2\\
    \label{proof equation G difference}
    \leq & 4(1+\rho ^2)L^2\mathbb{E}\left\|\*X_{k}-\overline{\*X}_{k}\*{1}_n^\top\right\|_F^2 + 4\alpha^2\rho ^2L^2\mathbb{E}\left\|\*Y_{k}-\overline{\*Y}_{k}\*{1}_n^\top\right\|_F^2 + 4\alpha^2nL^2\mathbb{E}\left\|\overline{\*Y}_{k}\right\|^2.
\end{align}

Putting it back we obtain
\begin{align}
    & \mathbb{E}\left\|\*Y_{k+1}-\overline{\*Y}_{k+1}\*{1}_n^\top\right\|_F^2\\
    \leq & \left(\frac{2\rho ^2}{1+\rho ^2}+\frac{16\alpha^2\rho ^4L^2}{1-\rho ^2}\right)\mathbb{E}\left\|\*Y_{k}-\overline{\*Y}_{k}\*{1}_n^\top\right\|_F^2 + \frac{16\rho ^2(1+\rho ^2)L^2}{1-\rho ^2}\mathbb{E}\left\|\*X_{k}-\overline{\*X}_k\*{1}_n^\top\right\|_F^2 +\frac{16\alpha^2\rho ^2nL^2}{1-\rho ^2}\mathbb{E}\left\|\overline{\*Y}_k\right\|^2\\
        & \frac{4\rho ^2}{1-\rho ^2}\mathbb{E}\left\|\*G_{k+2}-\*G_{k+1}-\*G_{k}+\*G_{k-1}\right\|_F^2 + 8n\rho ^2\tilde{\sigma}^2.
\end{align}
Combining Equation~(\ref{proof equation X-Xbar}) and Equation~(\ref{proof equation Y-Ybar}), we have
\begin{align}
    \begin{bmatrix}
    \mathbb{E}\left\|\*X_{k+1}-\overline{\*X}_{k+1}\*{1}_n^\top\right\|_F^2 \\
    \mathbb{E}\left\|\*Y_{k+1}-\overline{\*Y}_{k+1}\*{1}_n^\top\right\|_F^2
    \end{bmatrix}
    \preceq& 
    \begin{bmatrix}
    \*P_{11} & \*P_{12}\\
    \*P_{21} & \*P_{22}
    \end{bmatrix}
    \begin{bmatrix}
    \mathbb{E}\left\|\*X_{k}-\overline{\*X}_{k}\*{1}_n^\top\right\|_F^2 \\
    \mathbb{E}\left\|\*Y_{k}-\overline{\*Y}_{k}\*{1}_n^\top\right\|_F^2
    \end{bmatrix}\\
    & +
    \begin{bmatrix}
    0 \\
    \frac{4\rho ^2}{1-\rho ^2}\mathbb{E}\left\|{\*U_k}\right\|_F^2 + \frac{16\alpha^2\rho ^2nL^2}{1-\rho ^2}\mathbb{E}\left\|\overline{\*Y}_k\right\|^2 +  8n\rho ^2\tilde{\sigma}^2
    \end{bmatrix},
\end{align}
where
\begin{align}
    \*P_{11} = & \frac{2\rho ^2}{(1+\rho ^2)}\\
    \*P_{12} = & \frac{2\rho ^2\alpha^2}{1-\rho ^2}\\
    \*P_{21} = & \frac{16\rho ^2(1+\rho ^2)L^2}{1-\rho ^2}\\
    \*P_{22} = & \frac{2\rho ^2}{1+\rho ^2}+\frac{16\alpha^2\rho ^4L^2}{1-\rho ^2}\\
    \*U_k = & \*G_{k+2}-\*G_{k+1}-\*G_{k}+\*G_{k-1}.
\end{align}
For simplicity, define
\begin{align}
    \*z_k = & 
    \begin{bmatrix}
    \mathbb{E}\left\|\*X_{k+1}-\overline{\*X}_{k+1}\*{1}_n^\top\right\|_F^2 \\
    \mathbb{E}\left\|\*Y_{k+1}-\overline{\*Y}_{k+1}\*{1}_n^\top\right\|_F^2
    \end{bmatrix}\\
    \*P = & 
    \begin{bmatrix}
    \*P_{11} & \*P_{12}\\
    \*P_{21} & \*P_{22}
    \end{bmatrix}\\
    \*u_k = & 
    \begin{bmatrix}
    0 \\
    \frac{4\rho ^2}{1-\rho ^2}\mathbb{E}\left\|{\*U_k}\right\|_F^2 + \frac{16\alpha^2\rho ^2nL^2}{1-\rho ^2}\mathbb{E}\left\|\overline{\*Y}_k\right\|^2 +  8n\rho ^2\tilde{\sigma}^2,
    \end{bmatrix}
\end{align}
then we can write this linear system as
\begin{align}
    \*z_{k} \preceq \*P\*z_{k-1}+\*u_{k-1}\preceq \*P^k\*z_0 + \sum_{t=0}^{k-1}\*P^{k-t}\*u_t,
\end{align}
for simplicity. 

Let $\lambda_{1}(\*P), \lambda_{2}(\*P)$ denote the two eigenvalues of $\*P$ (without the loss of generality, we denote $\lambda_{1}(\*P)<\lambda_{2}(\*P)$),
define 
\begin{align}
\Psi=\sqrt{(\*P_{11}-\*P_{22})^2+4\*P_{12}\*P_{21}},
\end{align}
then with eigendecomposition, we obtain
\begin{align}
    \lambda_{1}(\*P) = & \frac{\*P_{11}+\*P_{22}-\Psi}{2}\\
    \lambda_{2}(\*P) = & \frac{\*P_{11}+\*P_{22}+\Psi}{2} = \frac{2\rho ^2}{1+\rho ^2}+\frac{8\alpha^2\rho ^4L^2}{1-\rho ^2} + \frac{16\alpha\rho ^2L\sqrt{\alpha^2\rho ^4L^2+(1+\rho ^2)}}{1-\rho ^2}\\
    \*P^k \preceq &
    \begin{bmatrix}
    \frac{\lambda_{1}^k(\*P)+\lambda_{2}^k(\*P)}{2}+\frac{(\*P_{11}-\*P_{22})(\lambda_{2}^k(\*P)-\lambda_{1}^k(\*P))}{2\Psi} & \frac{\*P_{12}}{\Psi}(\lambda_{2}^k(\*P)-\lambda_{1}^k(\*P))\\
    \frac{\*P_{21}}{\Psi}(\lambda_{2}^k(\*P)-\lambda_{1}^k(\*P)) & \frac{\lambda_{1}^k(\*P)+\lambda_{2}^k(\*P)}{2}+\frac{(\*P_{11}-\*P_{22})(\lambda_{1}^k(\*P)-\lambda_{2}^k(\*P))}{2\Psi}
    \end{bmatrix},
\end{align}
when the step size is small enough such that
\begin{equation}\label{proof equation step size requirement 1}
    \alpha L<\frac{(1-\rho )^2}{32},
\end{equation}
it can be verified that $\lambda_2(\*P)\leq\frac{\sqrt{\rho }+\rho }{1+\rho }$, and then we can compute the $\mathbb{E}\|\*X_k-\overline{\*X}_k\*1_n^\top\|^2$ and $\mathbb{E}\|\*Y_k-\overline{\*Y}_k\*1_n^\top\|^2$. We use $\*X[1:]$ to denote the first row of matrix $\*X$. First for $\*X_k$, we obtain:
\begin{align}
    \*P^k\*z_0[1:] \leq \*P_{12}k\lambda_2^{k-1}(\*P)\mathbb{E}\|\*Y_{0}-\overline{\*Y}_{0}\*{1}_n^\top\|_F^2 = \frac{2\rho ^2\alpha^2k}{1-\rho ^2}\lambda_2^{k-1}(\*P)\mathbb{E}\|\*Y_{0}-\overline{\*Y}_{0}\*{1}_n^\top\|_F^2
\end{align}
where we use the property that $\lambda_2^k(\*P)-\lambda_1^k(\*P)=(\lambda_2(\*P)-\lambda_1(\*P))\sum_{l=0}^{k-1}\lambda_2(\*P)^l\lambda_1(\*P)^{k-1-l}=\Psi k\lambda_2^{k-1}(\*P)$ and, 
similarly
\begin{align}
    & \*P^{k-t}\*u_t[1:]\\
        \leq &  \frac{2\rho ^2\alpha^2(k-t)}{1-\rho ^2}\lambda_2^{k-t-1}(\*P)\left(\frac{4\rho ^2}{1-\rho ^2}\mathbb{E}\left\|\*U_t\right\|_F^2 + \frac{16\alpha^2\rho ^2nL^2}{1-\rho ^2}\mathbb{E}\left\|\overline{\*Y}_t\right\|^2 +  8n\rho ^2\tilde{\sigma}^2\right)\\
    = &  \frac{2\rho ^2\alpha^2(k-t)}{1-\rho ^2}\lambda_2^{k-t-1}(\*P)\left(\frac{4\rho ^2}{1-\rho ^2}\mathbb{E}\left\|\*U_t\right\|_F^2 + \frac{16\alpha^2\rho ^2nL^2}{1-\rho ^2}\mathbb{E}\left\|\overline{\tilde{\*G}}_{t-1}\right\|^2 +  8n\rho ^2\tilde{\sigma}^2\right)\\
        \overset{(\ref{proof equation sampling noise})}{\leq} &  \frac{2\rho ^2\alpha^2(k-t)}{1-\rho ^2}\lambda_2^{k-t-1}(\*P)\left(\frac{4\rho ^2}{1-\rho ^2}\mathbb{E}\left\|\*U_t\right\|_F^2 + \frac{16\alpha^2\rho ^2nL^2}{1-\rho ^2}\mathbb{E}\left\|\overline{\*G}_{t-1}\right\|^2 + \frac{16\alpha^2\rho ^2\tilde{\sigma}^2L^2}{1-\rho ^2} +   8n\rho ^2\tilde{\sigma}^2\right),
\end{align}
then we obtain
\begin{align}
    & \mathbb{E}\left\|\*X_{k}-\overline{\*X}_{k}\*{1}_n^\top\right\|_F^2\\
    \leq & \frac{2\rho ^2\alpha^2k}{1-\rho ^2}\lambda_2^{k-1}(\*P)\mathbb{E}\|\*Y_{0}-\overline{\*Y}_{0}\*{1}_n^\top\|_F^2\\
        & + \sum_{t=0}^{k-1}\frac{2\rho ^2\alpha^2(k-t)}{1-\rho ^2}\lambda_2^{k-t-1}(\*P)\left(\frac{4\rho ^2}{1-\rho ^2}\mathbb{E}\left\|\*U_t\right\|_F^2 + \frac{16\alpha^2\rho ^2nL^2}{1-\rho ^2}\mathbb{E}\left\|\overline{\*G}_{t-1}\right\|^2 + \frac{16\alpha^2\rho ^2\tilde{\sigma}^2L^2}{1-\rho ^2} +   8n\rho ^2\tilde{\sigma}^2\right).
\end{align}
Summing over $k=0$ to $K-1$ we obtain
\begin{align}\label{proof equation sum Xk}
    & \sum_{k=0}^{K-1}\mathbb{E}\left\|\*X_{k}-\overline{\*X}_{k}\*{1}_n^\top\right\|_F^2\\
    \leq & \frac{2\rho ^2\alpha^2}{(1-\rho ^2)(1-\lambda_2(\*P))^2}\sum_{k=0}^{K-1}\mathbb{E}\|\*Y_{0}-\overline{\*Y}_{0}\*{1}_n^\top\|_F^2\\
        & + \frac{2\rho ^2\alpha^2}{(1-\rho ^2)(1-\lambda_2(\*P))^2}\sum_{k=0}^{K-1}\left(\frac{4\rho ^2}{1-\rho ^2}\mathbb{E}\left\|\*U_k\right\|_F^2 + \frac{16\alpha^2\rho ^2nL^2}{1-\rho ^2}\mathbb{E}\left\|\overline{\*G}_k\right\|^2 + \frac{16\alpha^2\rho ^2\tilde{\sigma}^2L^2}{1-\rho ^2} +   8n\rho ^2\tilde{\sigma}^2\right)\\
    \leq & \frac{2\rho ^2\alpha^2(1+\rho )nK\varsigma_0^2}{(1-\rho )(1-\sqrt{\rho })^2} + \frac{32\rho ^4\alpha^4nL^2}{(1-\rho )^2(1-\sqrt{\rho })^2}\sum_{k=0}^{K-1}\mathbb{E}\left\|\overline{\*G}_k\right\|^2 + \frac{8\rho ^4\alpha^2}{(1-\rho )^2(1-\sqrt{\rho })^2}\sum_{k=0}^{K-1}\mathbb{E}\|\*U_k\|_F^2\\
    & + \frac{32\rho ^4\alpha^4\tilde{\sigma}^2L^2K}{(1-\rho )^2(1-\sqrt{\rho })^2} +  \frac{8\rho ^4\alpha^2n\tilde{\sigma}^2(1+\rho )K}{(1-\rho )(1-\sqrt{\rho })^2}]\\
        \leq & \frac{2\rho ^2\alpha^2(1+\rho )nK\varsigma_0^2}{(1-\rho )(1-\sqrt{\rho })^2} + \frac{32\rho ^4\alpha^4nL^2}{(1-\rho )^2(1-\sqrt{\rho })^2}\sum_{k=0}^{K-1}\mathbb{E}\left\|\overline{\*G}_k\right\|^2 + \frac{8\rho ^4\alpha^2}{(1-\rho )^2(1-\sqrt{\rho })^2}\sum_{k=0}^{K-1}\mathbb{E}\|\*U_k\|_F^2\\
    & + \frac{16\rho ^4\alpha^2n\tilde{\sigma}^2(1+\rho )K}{(1-\rho )(1-\sqrt{\rho })^2},
\end{align}
where in the second step we used $\frac{1}{1-\lambda_2(\*P)}<\frac{1+\rho }{1-\sqrt{\rho }}$ since $\lambda_2(\*P)\leq\frac{\sqrt{\rho}+\rho}{1+\rho}$. And the third step holds due to Equation~(\ref{proof equation step size requirement 1}).
We proceed to analyze the case in $\*Y_k$:
we first have
\begin{align}
    [\*P^k]_{22} = & \frac{\lambda_{1}^k(\*P)+\lambda_{2}^k(\*P)}{2}+\frac{(\*P_{11}-\*P_{22})(\lambda_{1}^k(\*P)-\lambda_{2}^k(\*P))}{2\Psi}\\
    \leq & \lambda_2^k(\*P) + \frac{8\alpha^2\rho ^4L^2k\lambda_2^{k-1}(\*P)}{1-\rho ^2},
\end{align}
then we can have
\begin{align}
    \*P^k\*z_0[2:] \leq \left(\lambda_2^k(\*P) + \frac{8\alpha^2\rho ^4L^2k\lambda_2^{k-1}(\*P)}{1-\rho ^2}\right)\mathbb{E}\|\*Y_{0}-\overline{\*Y}_{0}\*{1}_n^\top\|_F^2,
\end{align}
and
\begin{align}
    & \*P^{k-t}\*u_t[2:]\\
    \leq &  \left(\lambda_2^{k-t}(\*P) + \frac{8\alpha^2\rho ^4L^2(k-t)\lambda_2^{k-t-1}(\*P)}{1-\rho ^2}\right)\cdot\left(\frac{4\rho ^2}{1-\rho ^2}\mathbb{E}\left\|\*U_t\right\|_F^2 + \frac{16\alpha^2\rho ^2nL^2}{1-\rho ^2}\mathbb{E}\left\|\overline{\*Y}_t\right\|^2 +  8n\rho ^2\tilde{\sigma}^2\right)\\
        \leq &  \left(\lambda_2^{k-t}(\*P) + \frac{8\alpha^2\rho ^4L^2(k-t)\lambda_2^{k-t-1}(\*P)}{1-\rho ^2}\right)\cdot\left(\frac{4\rho ^2}{1-\rho ^2}\mathbb{E}\left\|\*U_t\right\|_F^2 + \frac{16\alpha^2\rho ^2nL^2}{1-\rho ^2}\mathbb{E}\left\|\overline{\*G}_{t-1}\right\|^2 + \frac{16\alpha^2\rho ^2\tilde{\sigma}^2L^2}{1-\rho ^2} +   8n\rho ^2\tilde{\sigma}^2\right).
\end{align}
Summing over $k=0$ to $K-1$, we obtain
\begin{align}\label{proof equation sum Yk}
    & \sum_{k=0}^{K-1}\mathbb{E}\|\*Y_k-\overline{\*Y}_k\*1_n^\top\|^2\\
    \leq & \frac{(1+\rho )nK\varsigma_0^2}{1-\sqrt{\rho }} + \frac{8\alpha^2\rho ^4(1+\rho )L^2nK\varsigma_0^2}{(1-\rho )(1-\sqrt{\rho })^2}\\
        & + \left(\frac{1+\rho }{1-\sqrt{\rho }} + \frac{8\alpha^2\rho ^4(1+\rho )L^2}{(1-\rho )(1-\sqrt{\rho })^2}\right)\sum_{k=0}^{K-1}\left(\frac{4\rho ^2}{1-\rho ^2}\mathbb{E}\left\|\*U_k\right\|_F^2 + \frac{16\alpha^2\rho ^2nL^2}{1-\rho ^2}\mathbb{E}\left\|\overline{\*G}_{k}\right\|^2 + \frac{16\alpha^2\rho ^2\tilde{\sigma}^2L^2}{1-\rho ^2} +   8n\rho ^2\tilde{\sigma}^2\right).
\end{align}
We next solve $\|\*U_k\|_F$, from the definition of $\*U_k$ we obtain that
\begin{align}
    \sum_{k=0}^{K-1}\mathbb{E}\left\|\*U_k\right\|_F^2 = & \sum_{k=0}^{K-1}\mathbb{E}\left\|\*G_{k+2}-\*G_{k+1}-\*G_{k}+\*G_{k-1}\right\|_F^2\\
        \leq & 2\sum_{k=0}^{K-1}\mathbb{E}\left\|\*G_{k+2}-\*G_{k+1}\right\|_F^2 + 2\sum_{k=0}^{K-1}\mathbb{E}\left\|\*G_{k}-\*G_{k-1}\right\|_F^2\\
    \leq & 4\sum_{k=0}^{K-1}\mathbb{E}\left\|\*G_{k+2}-\*G_{k+1}\right\|_F^2\\
    = & 4\sum_{k=0}^{K-1}\sum_{i=1}^{n}\mathbb{E}\left\|\nabla f(\*x_{k+1,i})-\nabla f(\*x_{k,i})\right\|^2\\
    \leq & 4L^2\sum_{k=0}^{K-1}\sum_{i=1}^{n}\mathbb{E}\left\|\*x_{k+1,i}-\*x_{k,i}\right\|^2\\
        = & 4L^2\sum_{k=0}^{K-1}\mathbb{E}\left\|\*X_{k+1}-\*X_{k}\right\|_F^2.
\end{align}
Fit in the derivation from Equation~(\ref{proof equation G difference}) we obtain
\begin{align}
    &\sum_{k=0}^{K-1}\mathbb{E}\left\|\*U_k\right\|_F^2\\
    \leq & 4L^2\sum_{k=0}^{K-1}\mathbb{E}\left\|\*X_{k+1}-\*X_k\right\|_F^2\\
    \leq & 16(1+\rho ^2)L^2\sum_{k=0}^{K-1}\mathbb{E}\left\|\*X_{k}-\overline{\*X}_{k}\*{1}_n^\top\right\|_F^2 + 16\alpha^2\rho ^2L^2\sum_{k=0}^{K-1}\mathbb{E}\left\|\*Y_{k}-\overline{\*Y}_{k}\*{1}_n^\top\right\|_F^2 + 16\alpha^2nL^2\sum_{k=0}^{K-1}\mathbb{E}\left\|\overline{\*Y}_{k}\right\|^2\\
        \leq & \frac{32\rho ^2\alpha^2(1+\rho )^2nK\varsigma_0^2L^2}{(1-\rho )(1-\sqrt{\rho })^2}
    + \frac{512\rho ^4(1+\rho )\alpha^4nL^4}{(1-\rho )^2(1-\sqrt{\rho })^2}\sum_{k=0}^{K-1}\mathbb{E}\left\|\overline{\*G}_k\right\|^2
    + \frac{256\rho ^4(1+\rho )\alpha^2L^2}{(1-\rho )^2(1-\sqrt{\rho })^2}\sum_{k=0}^{K-1}\mathbb{E}\|\*U_k\|_F^2\\
    &
    +  \frac{256\rho ^4\alpha^2n\tilde{\sigma}^2(1+\rho )^2KL^2}{(1-\rho )(1-\sqrt{\rho })^2}\\
        & + \frac{16\alpha^2\rho ^2(1+\rho )nK\varsigma_0^2L^2}{1-\sqrt{\rho }} + \frac{128\alpha^4\rho ^6(1+\rho )L^4nK\varsigma_0^2}{(1-\rho )(1-\sqrt{\rho })^2}\\
    & + \left(\frac{16\alpha^2\rho ^2(1+\rho )L^2}{1-\sqrt{\rho }} + \frac{128\alpha^4\rho ^6(1+\rho )L^4}{(1-\rho )(1-\sqrt{\rho })^2}\right)\sum_{k=0}^{K-1}\frac{4\rho ^2}{1-\rho ^2}\mathbb{E}\left\|\*U_k\right\|_F^2\\
        & +  \left(\frac{16\alpha^2\rho ^2(1+\rho )L^2}{1-\sqrt{\rho }} + \frac{128\alpha^4\rho ^6(1+\rho )L^4}{(1-\rho )(1-\sqrt{\rho })^2}\right)\sum_{k=0}^{K-1}\left( \frac{16\alpha^2\rho ^2nL^2}{1-\rho ^2}\mathbb{E}\left\|\overline{\*G}_{k}\right\|^2 + \frac{16\alpha^2\rho ^2\tilde{\sigma}^2L^2}{1-\rho ^2} +   8n\rho ^2\tilde{\sigma}^2\right)\\
    & + 16\alpha^2nL^2\sum_{k=0}^{K-1}\mathbb{E}\left\|\overline{\*Y}_{k}\right\|^2\\
    \leq & \frac{64\rho ^2\alpha^2(1+\rho )^2nK\varsigma_0^2L^2}{(1-\rho )(1-\sqrt{\rho })^2}
    + \frac{512\rho ^4(1+\rho )\alpha^2L^2}{(1-\rho )^2(1-\sqrt{\rho })^2}\sum_{k=0}^{K-1}\mathbb{E}\|\*U_k\|_F^2 + \frac{512\rho ^4\alpha^2n\tilde{\sigma}^2(1+\rho )^2KL^2}{(1-\rho )(1-\sqrt{\rho })^2}\\
    & + 32\alpha^2nL^2\sum_{k=0}^{K-1}\mathbb{E}\left\|\overline{\*G}_{k}\right\|^2,
\end{align}
where in the third step we use the derivation from Equation~(\ref{proof equation sum Xk}) and (\ref{proof equation sum Yk}), in the fourth step we repeatedly use Equation~(\ref{proof equation step size requirement 1}) and Equation~(\ref{proof equation sampling noise}),
solve it we obtain
\begin{align}
    \sum_{k=0}^{K-1}\mathbb{E}\left\|\*U_k\right\|_F^2\leq \frac{128\rho ^2\alpha^2(1+\rho )^2nK\varsigma_0^2L^2}{(1-\rho )(1-\sqrt{\rho })^2}
    + \frac{1024\rho ^4\alpha^2n\tilde{\sigma}^2(1+\rho )^2KL^2}{(1-\rho )(1-\sqrt{\rho })^2} + 64\alpha^2nL^2\sum_{k=0}^{K-1}\mathbb{E}\left\|\overline{\*G}_{k}\right\|^2,
\end{align}
where again we use Equation~(\ref{proof equation step size requirement 1}),
combine it with Equation~(\ref{proof equation sum Xk}) we obtain
\begin{align}
    & \sum_{k=0}^{K-1}\mathbb{E}\left\|\*X_{k}-\overline{\*X}_{k}\*{1}_n^\top\right\|_F^2\\
    \label{proof equation final X-Xbar}
        \leq & \frac{4\rho ^2\alpha^2(1+\rho )nK\varsigma_0^2}{(1-\rho )(1-\sqrt{\rho })^2} + \frac{544\rho ^4\alpha^4nL^2}{(1-\rho )^2(1-\sqrt{\rho })^2}\sum_{k=0}^{K-1}\mathbb{E}\left\|\overline{\*G}_k\right\|^2 + \frac{32\rho ^4\alpha^2n\tilde{\sigma}^2(1+\rho )K}{(1-\rho )(1-\sqrt{\rho })^2},
\end{align}
where we use Equation~(\ref{proof equation step size requirement 1}). 

Recall from Equation~(\ref{proof equation main formula}) that
\begin{align}
    & \sum_{k=0}^{K-1}\alpha(1-\alpha L-24\alpha^2L^2)\left\|\overline{\*G}_k\right\|^2 + \sum_{k=0}^{K-1}\alpha\mathbb{E}\left\|\nabla f\left(\overline{\*X}_k\right)\right\|^2\\
    \leq & 2\Delta + \frac{\alpha^2\tilde{\sigma}^2LK}{n} + \frac{16\alpha L^2}{n}\sum_{k=0}^{K-1}\mathbb{E}\left\|\*X_k-\overline{\*X}_k\*{1}_n^\top\right\|_F^2 + \frac{24\alpha^3\tilde{\sigma}^2L^2K}{n}.
\end{align}
Combine Equation~(\ref{proof equation step size requirement 1}) and (\ref{proof equation final X-Xbar}),
we obtain
\begin{align}
    \frac{1}{K}\sum_{k=0}^{K-1}\mathbb{E}\left\|\nabla f\left(\overline{\*X}_k\right)\right\|^2
        \leq O\left(\frac{\Delta}{\alpha K} + \frac{\alpha\tilde{\sigma}^2L}{n} + \frac{\rho ^2\alpha^2L^2\varsigma_0^2}{(1-\rho )^3}+ \frac{\rho ^4\alpha^2\tilde{\sigma}^2L^2}{(1-\rho )^3} + \frac{\alpha^2\tilde{\sigma}^2L^2}{n}\right),
\end{align}
where we omit the numerical constants. Set 
\begin{align}
    \alpha=\frac{1}{\tilde{\sigma}\sqrt{KL/n\Delta}+\frac{\rho ^{\frac{2}{3}}L^{\frac{2}{3}}\varsigma_0^{\frac{2}{3}}K^\frac{1}{3}}{\Delta^{\frac{1}{3}}(1-\rho )}+\frac{32L}{(1-\rho )^2}},
\end{align}
we obtain
\begin{align}
    \frac{1}{K}\sum_{k=0}^{K-1}\mathbb{E}\left\|\nabla f\left(\overline{\*X}_k\right)\right\|^2 \leq O\left(\frac{\sqrt{\Delta L}\tilde{\sigma}}{\sqrt{nK}} + \frac{(\rho \Delta L\varsigma_0)^{\frac{2}{3}}}{(1-\rho )K^{\frac{2}{3}}} +  \frac{\rho ^2n\Delta L}{(1-\rho )^3K} + \frac{\Delta L}{(1-\rho )^2K}\right).
\end{align}
Fit in $T=KR$ and $\tilde{\sigma}^2=\sigma^2/BR$, we obtain
\begin{align}
    \frac{1}{K}\sum_{k=0}^{K-1}\mathbb{E}\left\|\nabla f\left(\overline{\*X}_k\right)\right\|^2\leq O\left(\frac{\sqrt{\Delta L}\sigma}{\sqrt{nBT}} + \frac{(\rho \Delta L\varsigma_0R)^{\frac{2}{3}}}{(1-\rho )T^{\frac{2}{3}}} + \frac{\rho ^2nR\Delta L}{(1-\rho )^3T} + \frac{R\Delta L}{(1-\rho )^2T} \right),
\end{align}
set 
\[
R 
= 
\frac{1}{\sqrt{1-\lambda_2(\*W)}}
\max\left( 
    \frac{1}{2} \log(n),
    \frac{1}{2} \log\left( \frac{\varsigma_0^2 T}{\Delta L} \right)
\right),
\]
we first have $\rho\leq 1/\sqrt{2}$ since
\begin{align}
    R\geq\frac{\log(n)}{2\sqrt{1-\lambda_2(\*W)}} \geq \frac{-\log(n)}{2\log(1-\sqrt{1-\lambda_2(\*W)})}
    \Rightarrow \left(1-\sqrt{1-\lambda_2(\*W)}\right)^{R}\leq \frac{1}{\sqrt{n}} \leq \frac{1}{\sqrt{2}},
\end{align}
this implies
\begin{align}
    \frac{1}{K}\sum_{k=0}^{K-1}\mathbb{E}\left\|\nabla f\left(\overline{\*X}_k\right)\right\|^2 &\leq O\left(\frac{\sqrt{\Delta L}\sigma}{\sqrt{nBT}} + \frac{(\rho \Delta L\varsigma_0R)^{\frac{2}{3}}}{T^{\frac{2}{3}}} + \frac{\rho ^2nR\Delta L}{T} + \frac{R\Delta L}{T} \right)
    \\&\leq O\left(\frac{\sqrt{\Delta L}\sigma}{\sqrt{nBT}} + \frac{(\rho  \varsigma_0 \sqrt{T} R / \sqrt{\Delta L})^{\frac{2}{3}} \Delta L}{T} + \frac{\rho ^2nR\Delta L}{T} + \frac{R\Delta L}{T} \right),
\end{align}
with the assignment of $R$,
$\rho^2 n < 1$ and $\rho  \varsigma_0 \sqrt{T} / \sqrt{\Delta L} < 1$, so since it also holds that $R \ge 1$ (and so $R^{2/3} \le R$),
\begin{align}
    \min_t\|\nabla f\left(\overline{\*X}_t\right)\|^2 \leq & \frac{1}{K}\sum_{k=0}^{K-1}\mathbb{E}\left\|\nabla f\left(\overline{\*X}_k\right)\right\|^2\\
    \leq & O\left(\frac{\sqrt{\Delta L}\sigma}{\sqrt{nBT}} + \frac{(R)^{\frac{2}{3}} \Delta L}{T} + \frac{R\Delta L}{T} + \frac{R\Delta L}{T} \right) 
    \\ \leq & O\left(\frac{\sqrt{\Delta L}\sigma}{\sqrt{nBT}} + \frac{R\Delta L}{T} \right)
    \\= & O\left(\frac{\sqrt{\Delta L}\sigma}{\sqrt{nBT}} + \frac{\Delta L}{T \sqrt{1-\lambda_2(\*W)}} \cdot
    \max\left( 
    \log(n),
    \log\left( \frac{\varsigma_0^2 T}{\Delta L} \right) \right) \right),
\end{align}
when
\begin{align}
    T \leq O\left(\frac{\Delta L\sigma^2}{nB\epsilon^4}\right),
\end{align}
we have
\begin{align}
    \frac{\sqrt{\Delta L}\sigma}{\sqrt{nBT}}\leq O(\epsilon^2).
\end{align}
On the other hand,
when
\begin{align}
    T \leq O\left(\max\left(\frac{\log(n)\Delta L}{\epsilon^2\sqrt{1-\lambda_2(\*W)}}, \frac{\Delta L}{\epsilon^2\sqrt{1-\lambda_2(\*W)}}\log\left(\frac{\varsigma_0^2}{\epsilon^2\Delta L}\right)\right) \right),
\end{align}
we have
\begin{align}
    \frac{\Delta L}{T \sqrt{1-\lambda_2(\*W)}} \cdot
    \max\left( 
    \log(n),
    \log\left( \frac{\varsigma_0^2 T}{\Delta L} \right) \right)\leq O(\epsilon^2),
\end{align}
to see this, note that
\begin{align}
    \frac{\Delta L}{T \sqrt{1-\lambda_2(\*W)}}
    \log\left( \frac{\varsigma_0^2 T}{\Delta L} \right)=& \epsilon^2\frac{\log\left( \frac{\varsigma_0^2}{\epsilon^2\Delta L} \log\left( \frac{\varsigma_0^2}{\epsilon^2\Delta L} \right)\right)}{\log\left( \frac{\varsigma_0^2 }{\epsilon^2\Delta L} \right)} \leq O(\epsilon^2).
\end{align}
Finally, we can obtain the upper bound
\begin{align}
    T \leq & O\left(\frac{\Delta L\sigma^2}{nB\epsilon^4} + \max\left(\frac{\log(n)\Delta L}{\epsilon^2\sqrt{1-\lambda_2(\*W)}}, \frac{\Delta L}{\epsilon^2\sqrt{1-\lambda_2(\*W)}}\log\left(\frac{\varsigma_0^2}{\epsilon^2\Delta L}\right)\right)\right)\\
    = & O\left(\frac{\Delta L\sigma^2}{nB\epsilon^4} + \frac{\Delta L}{\epsilon^2\sqrt{1-\lambda_2(\*W)}}\log\left(n + \frac{\varsigma_0n}{\epsilon\sqrt{\Delta L}}\right)\right),
\end{align}
as desired.
\end{proof}

\section{Details to footnotes}\label{section footnote}
\subsection{Asynchronous Algorithm (Footnote 2)}
In the full paper, we focus on the synchronous algorithms, i.e., we assume the existence of a synchronization process among workers between two adjacent iterations. 
We now extend our formulation to asynchronous algorithms. Since workers now update and communicate asynchronously, we define any gradient update that took place on a randomly chosen worker as one iteration. This randomness depends on system implementation, stochastic events, etc. This is a commonly adopted definition in the analysis of (decentralized) asynchronous algorithms \cite{lian2017asynchronous}.
To obtain a lower bound in such case, consider the two settings as shown in the proof of Theorem~\ref{thmlowerbound}. In setting 1, it can be easily verified that the lower bound for sample complexity is
\begin{equation}
    \Omega\left(\frac{\Delta L\sigma^2}{B\epsilon^4}\right).
\end{equation}
This holds because in the extreme case, only one worker is making contributions to the optimization. And since we have not made any assumption on how workers are sampled to conduct the next iteration, this is a valid bound for arbitrary distribution.
On the other hand, considering setting 2, the lower bound is still $\Omega(T_0D)$ where $T_0=\Omega(\Delta L\epsilon^{-2})$ is the lower bound in the sequential case, since the systems need at least $\Omega(D)$ iterations for the workers in $I_0$ and $I_2$ to contact. The lower bound for communication complexity is then
\begin{equation}
    \Omega\left(\frac{\Delta LD}{\epsilon^2}\right).
\end{equation}
Combining them together, we can get the final lower bound as:
\begin{equation}
    \Omega\left(\frac{\Delta L\sigma^2}{B\epsilon^4} + \frac{\Delta LD}{\epsilon^2}\right).
\end{equation}
Note that this bound holds with probability $1$. It is possible to propose finer-grained assumption on how workers are chosen (e.g. uniformly random) and use concentration inequalities (e.g. Hoeffding's inequality) to get tighter bounds, we leave this as future work.

\subsection{Relax zero-respecting assumption (Footnote 3)}
To relax the zero-respecting assumption, we can use the technique proposed by \cite{carmon2019lower} (See their proofs to Proposition 1 and 2). The basic idea is that to adversarially construct the loss function and rotate the non-zero coordinates in $t$-th iterations, such that when the algorithm operates on the rotated function, the first $t$ iterations match with that of the old function. However, the new rotated function is still zero-respecting to the algorithm after $t$-th iteration so is generally hard to optimize.
The details can be found in \cite{carmon2019lower}.

\subsection{Specific algorithm for Average Consensus (Footnote 8)}
Many algorithms have been proposed on solving the Average Consensus problem, readers can find details in many previous works on graph theory such as \cite{georgopoulos2011definitive,hendrickx2014graph,ko2010matrix}.
A straightforward algorithm is the Minimum Spanning Tree, that is, we first generate a spanning tree of the graph, and then the workers send and receive message using propagation on the tree. Specifically, starting from the leaves, all the children nodes of the tree send its accumulated value to the parents and the root compute the averaged value after gathering the information from the graph. And then reversely, the parent nodes send the value back to the child nodes and eventually all the nodes will get the averaged value. This algorithm is also known as the \texttt{GATHER-PROPAGATE} algorithm as discussed in \cite{ko2010matrix}, section 3. We include the detailed pseudo-code\footnote{This code is proposed by \citet{ko2010matrix}, we do not intend to take credit for this.} in Algorithm~\ref{Average Consensus example}.
\begin{algorithm}[t]
\small
	\caption{GATHER-PROPAGATE (Spanning Tree) for a single coordinate}\label{Average Consensus example}
	\begin{algorithmic}[1]
		\Require communication graph $G$, a single coordinate on workers (all the coordinates follow the same instructions) to be communicated $\*X\in\mathbb{R}^{n}$.
		
		\State $\*d\leftarrow$ vector of 1's indexed by $V(G)$ (vertices set of graph $G$).
		
		\State $\mathcal{I}\leftarrow$ a spnning tree of $G$ with root $r$ arbitrarily picked.
		
		\For{$v\in V(\mathcal{I})$}
		    \State $l_v\leftarrow \bar{D}(r, v)$ (the distance between $r$ and $v$)
		\EndFor
		
		\For{$\alpha=\max_vl_v, \cdots, 1$}
		\For{$v$ with $l_v=\alpha$}
		    \State $v$ gives all its value onto its parents $u$:
		    \begin{align*}
		        \begin{bmatrix}
                \*X_u\\
                \*X_v
                \end{bmatrix}
                \leftarrow
		        \begin{bmatrix}
                1 & 1\\
                0 & 0
                \end{bmatrix}
                \begin{bmatrix}
                \*X_u\\
                \*X_v
                \end{bmatrix}
		    \end{align*}
		    \State $\*d_u\leftarrow\*d_u+\*d_v$
		\EndFor
		\EndFor
		\For{$\alpha=0, \cdots, \max_vl_v-1$}
		\For{$u$ with $l_u=\alpha$}
		    \State $\{v_1, \cdots, v_\beta\}\leftarrow$ set of children of $u$
		    \State re-distribute the results:
		    \begin{align*}
		        \begin{bmatrix}
                \*X_u\\
                \*X_{v_1}\\
                \vdots\\
                \*X_{v_{\beta}}\\
                \end{bmatrix}
                \leftarrow\frac{1}{\*d_u}
		        \begin{bmatrix}
                \*d_u-\*d_u-\cdots -\*d_{v_\beta}\\
                \*d_{v_1}\\
                \vdots\\
                \*d_{v_{\beta}}
                \end{bmatrix}
                \*X_u
		    \end{align*}
		\EndFor
		\EndFor
		\State \textbf{return} $X\frac{\*1\*1^\top}{n}$
	\end{algorithmic}
\end{algorithm}

\end{document}